\documentclass[10pt,twocolumn,letterpaper]{article}

\usepackage{cvpr}              

\usepackage{array}

\newcommand{\LiRGB}{LiDAR\text{-}RGB\xspace}
\newcommand{\LiEvent}{LiDAR\text{-}Event\xspace}
\newcolumntype{L}[1]{>{\raggedright\arraybackslash}p{#1}} 
\newcolumntype{C}[1]{>{\centering\arraybackslash}p{#1}}   

\usepackage{amsmath}
\usepackage{booktabs}
\usepackage{amssymb}
\usepackage{float}
\usepackage{multirow}
\usepackage{algorithm}
\usepackage{algorithmic}
\usepackage{tabularx,booktabs,array}
\usepackage{tabularx}
\usepackage{array}
\usepackage{subcaption}
\usepackage{makecell}
\newcolumntype{Y}{>{\raggedright\arraybackslash}X}
\usepackage{siunitx} 
\newcommand{\cmark}{\ding{51}} 
\usepackage{pifont}
\usepackage[table]{xcolor}
\usepackage[percent]{overpic}

\definecolor{cvprblue}{rgb}{0.21,0.49,0.74}
\usepackage[pagebackref,breaklinks,colorlinks,allcolors=cvprblue]{hyperref}
\usepackage[nameinlink,sort&compress,capitalize]{cleveref}
\creflabelformat{equation}{#2#1#3}

\def\paperID{---} 
\def\confName{CVPR}
\def\confYear{2026}

\title{LiREC-Net: A Target-Free and Learning-Based Network\\for LiDAR, RGB, and Event Calibration}

\author{Aditya Ranjan Dash$^1$ \hspace{1.5em} Ramy Battrawy$^2$ \hspace{1.5em} René Schuster$^{1,2}$ \hspace{1.5em} Didier Stricker$^{1,2}$ \\
$^{1}$RPTU -- University Kaiserslautern-Landau\\
$^{2}$DFKI -- German Research Center for Artificial Intelligence\\
{\tt\small firstname.lastname@dfki.de}
}

\begin{document}

\maketitle

\begin{abstract}
Advanced autonomous systems rely on multi-sensor fusion for safer and more robust perception.
To enable effective fusion, calibrating directly from natural driving scenes (\ie, target-free) with high accuracy is crucial for precise multi-sensor alignment.
Existing learning-based calibration methods are typically designed for only a single pair of sensor modalities (\ie, a bi-modal setup).
Unlike these methods, we propose LiREC-Net, a target-free, learning-based calibration network that jointly calibrates multiple sensor modality pairs, including LiDAR, RGB, and event data, within a unified framework.
To reduce redundant computation and improve efficiency, we introduce a shared LiDAR representation that leverages features from both its 3D nature and projected depth map, ensuring better consistency across modalities.
Trained and evaluated on established datasets, such as KITTI and DSEC, our LiREC-Net achieves competitive performance to bi-modal models and sets a new strong baseline for the tri-modal use case.


\end{abstract}

\begin{figure}[t]
    \centering
    \includegraphics[width=\columnwidth]{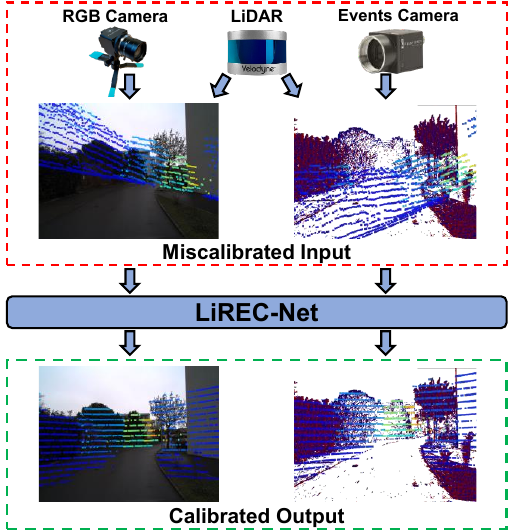}
    \caption{Our LiREC-Net takes miscalibrated tri-modal inputs and learns to produce spatially aligned outputs. The top row shows the raw miscalibrated overlays, while the bottom row illustrates the calibrated LiDAR projected onto both RGB and event frames.} 
    \label{fig:teaser}
\end{figure}

\section{Introduction} \label{sec:intro}
Modern vehicles with assisted and autonomous driving systems use multiple sensors together or fuse their outputs to build a consistent view of the environment.
Calibration is needed to relate the relative poses of all sensors in a common coordinate system and to keep their measurements aligned in space.
Sensors such as LiDAR, RGB cameras, and event cameras provide complementary information, and calibration ensures that this information corresponds spatially across views.
In real deployments, vehicles experience vibration, temperature change, minor impacts, and routine maintenance, and over time, these factors can shift sensor poses even when the initial setup was correct.
Traditional target-based methods, such as \cite{geiger, dhall2017lidarcameracalibrationusing3d3d, pandey2012automatic}, which use checkerboards or markers, can be accurate, but they require controlled space, careful placement, repeated captures, and human oversight, which interrupts operation and is expensive to perform frequently. 
A target-free procedure avoids special setups and can be repeated whenever needed. Calibrating directly in natural scenes removes the need for controlled environments and reflects real operating conditions.
Building on this idea, learning-based calibration methods leverage deep neural networks to extract and match cross-modal features from natural scenes.
Methods such as LCCNet \cite{LCCNet}, RegNet \cite{RegNet}, CalibNet \cite{CalibNet}, LCCRAFT \cite{LCCRAFT}, PseudoCal \cite{pseudocal}, and MULiEv \cite{muliev} eliminate the need for calibration targets, improve robustness under varying conditions, and enable automated, target-free calibration.
However, these methods are typically limited to only a single pair of sensor modalities (\ie, the bi-modal setup), either \LiRGB or \LiEvent, which duplicates effort and risks inconsistencies when extending to all three sensors.
Inspired by these learning-based methods, we aim to address this gap by developing a unified, tri-modal calibration framework that jointly handles LiDAR, RGB, and event cameras.
We propose LiREC-Net, an end-to-end learning-based calibration network that performs extrinsic calibration directly from natural driving scenes, as illustrated in \cref{fig:teaser}, without requiring special targets or controlled environments.
The same model can also operate in reduced pairwise configurations, while maintaining a consistent alignment path across all three modalities when available.

\noindent Our main contributions are:
\begin{itemize}
    \item LiREC-Net -- a unified and novel tri-modal neural network for calibrating LiDAR, RGB, and event cameras, improving both calibration accuracy and efficiency.
    \item A shared LiDAR representation used by both the \LiRGB and \LiEvent paths, improving consistency across pairs and reducing redundancy.
    \item A point-cloud encoding strategy that fuses the 3D structure with their projected depth maps, enhancing the overall point cloud representation and improving accuracy.
    \item A strong tri-modal baseline with consistent calibration results across datasets and competitive errors compared to established bi-modal methods.
\end{itemize}

\section{Related Work} \label{sec:related}
Calibration has been extensively explored in pairwise  (\ie, bi-modal) settings, encompassing target-based, target-free, and learning-based approaches. Below, we review prior work on \LiRGB and \LiEvent calibration.

\subsection{\LiRGB Calibration} \label{sec:related:camera-lidar}
Target-based approaches use multiple checkerboards to obtain sufficient constraints for full 6-DoF calibration \cite{geiger}. Likewise, Dhall \etal \cite{dhall2017lidarcameracalibrationusing3d3d} use ArUco markers for 3D–3D calibration with a fast closed-form extrinsic solution.
Beyond target-based approaches, target-free methods operate in natural scenes by maximizing mutual information between image intensity and LiDAR reflectance \cite{pandey2012automatic}.

Deep learning enables a new class of target-free calibration methods, in which a neural network is trained to align the sensors’ data.
RegNet \cite{RegNet} is one of the earliest models in this category, directly regressing extrinsic parameters from RGB images and 3D point clouds. 
CalibNet \cite{CalibNet} introduces a geometric, self-supervised approach using depth-photometric and point-cloud distance losses. 
LCCNet \cite{LCCNet} predicts extrinsics using a regression framework built on cost volumes constructed from image features and projected LiDAR-depth features, capturing cross-modal similarity.
LCCRAFT \cite{LCCRAFT} adapts RAFT \cite{teed2020raft} for calibration by iteratively updating the pose estimate, achieving improved robustness to large initial miscalibrations.
PseudoCal \cite{pseudocal}, which uses PointPillars \cite{pointpillars} for feature extraction, processes the RGB image with \cite{monoculardepth} to generate a 3D pseudo-LiDAR that is then used for calibration together with the real LiDAR.
To further improve accuracy, Cocheteux \etal \cite{unical} employ two cascaded refinement stages, UniCal-M and UniCal-S, yielding precise extrinsic calibration.

Our work draws inspiration from LCCNet \cite{LCCNet} and extends the core concept to a unified tri-modal setting.

\subsection{\LiEvent Calibration} \label{sec:related:Li-Event}
Similar to \LiRGB calibration, \LiEvent calibration has been explored with conventional methods. 
For example, L2E \cite{L2E} projects the LiDAR into a 2D image and recovers the pose by maximizing mutual information between projected LiDAR intensity and accumulated event frames. EF-Calib \cite{EF-Calib} jointly estimates intrinsics, extrinsics, and time offset for event and frame cameras using a custom calibration pattern and a continuous-time motion model. The method detects the shared pattern in both sensors, models motion with a piecewise B-spline, derives analytical Jacobians, and solves a spatiotemporal bundle-like optimization to recover all parameters jointly.

Target-free, learning-based \LiEvent calibration is scarcely explored, with MULiEv \cite{muliev} being the only representative method.
Using the DSEC dataset \cite{DSEC}, MULiEv concatenates the LiDAR depth image and event frame channel-wise, extracts features with MViTV2 \cite{mvitv2}, and predicts the extrinsics.
It also employs iterative refinement, similar to LCCNet \cite{LCCNet}, but with fewer stages.

In contrast to prior learning-based approaches that address only \LiRGB or \LiEvent calibration independently, our method jointly calibrates both pairs within a single target-free architecture. It reuses a shared LiDAR backbone with point-based and depth-based feature fusion, constructs pairwise cost volumes for each modality, and supports operation with either or both camera inputs.

\section{LiREC-Net} \label{sec:method}

\subsection{Problem Formulation} \label{sec:method:problem}
In this work, we address the joint calibration of a LiDAR sensor, an RGB camera, and an event camera.
Each training or evaluation sample consists of a miscalibrated LiDAR point cloud $P \in \mathbb{R}^{N\times 3}$, an RGB image $I \in \mathbb{R}^{H\times W\times 3}$, an event representation $E \in \mathbb{R}^{H\times W\times 2}$, and known camera intrinsics $\mathbf{K}_{\mathrm{RGB}}$ and $\mathbf{K}_{\mathrm{Ev}}$. 
The network takes these three modalities as input and predicts the LiDAR-to-camera extrinsics (\ie, relative poses) for multiple modality pairs $v \in \{\mathrm{Li}-\mathrm{RGB}, \mathrm{Li}-\mathrm{Ev}\}$.
These extrinsics are parameterized by a translation $\hat{\mathbf t}^{v} \in \mathbb{R}^3$ and a quaternion rotation $\hat{\mathbf q}^{v} \in \mathbb{R}^4$, yielding the predicted transformations $\hat{\mathbf T}^{v}$.
We assume that the transformations per pair are different from each other, while RGB and event cameras are assumed to be pre-calibrated, so the relative pose $\mathbf{T}^{\mathrm{Ev}\rightarrow\mathrm{RGB}}$ is known. 
When only one camera modality is available, the model predicts the corresponding single transformation.

\begin{figure*}[t]
    \centering
    \includegraphics[width=\textwidth]{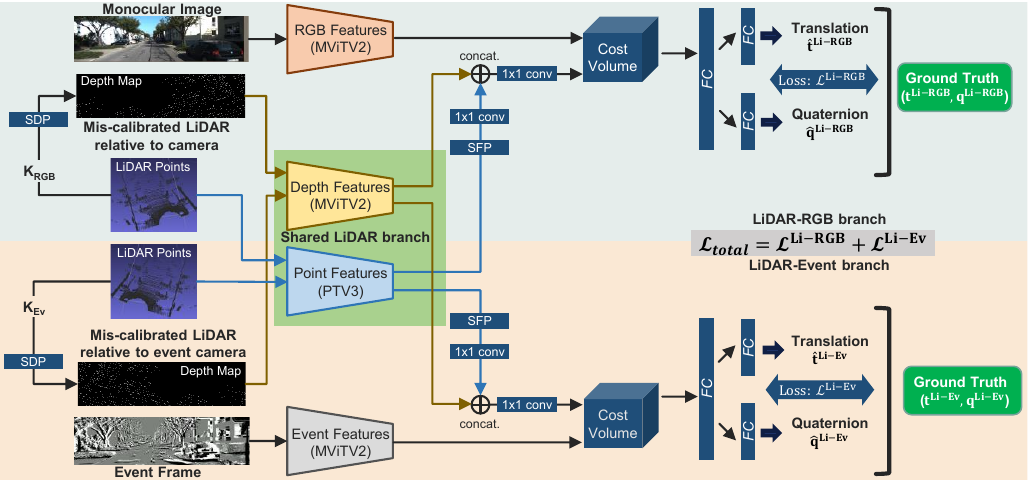}
    \caption{Overview of LiREC-Net. A miscalibrated LiDAR point cloud $P$ is processed by two LiDAR encoders (point- and depth-based). Point features are projected to the image plane using known intrinsics $\mathbf{K}_{\mathrm{RGB}}$ and $\mathbf{K}_{\mathrm{Ev}}$, and then \emph{fused} with depth features to form a unified LiDAR embedding. In parallel, the RGB image $I$ and event representation $E$ are encoded by their respective encoders. The unified LiDAR embedding is combined with the corresponding RGB/event features to build two pair-wise cost volumes, which are refined by context modules and passed to prediction heads that output the \LiRGB and \LiEvent extrinsics $\hat{\mathbf T}^{\mathrm{Li}-\mathrm{RGB}}$ and $\hat{\mathbf T}^{\mathrm{Li}-\mathrm{Ev}}$.}
    \label{fig:architecture}
\end{figure*}

\subsection{Architecture} \label{sec:method:architecture}
LiREC-Net follows a dual-path design that jointly handles \LiRGB and \LiEvent calibration within a unified framework.
The overall architecture, shown in \cref{fig:architecture}, consists of modality-specific feature encoders followed by pair-wise cost volumes, context modules, and prediction heads.

\subsubsection{Input Processing} \label{sec:method:architecture:preprocess}
As LiDAR, RGB, and event cameras are different modalities, we preprocess them separately before encoding. All three sensors need to be temporally synchronized before preprocessing. The following outlines the preprocessing steps for each.

\paragraph{LiDAR.}
For the LiDAR input, we first transform all points into a camera coordinate system, remove those with negative depth, and apply a cutoff at a maximum range $Z_{\max}$. The resulting point cloud is then resampled to a fixed number of $N$ points. 
Frames with more than $N$ points are randomly downsampled, while frames with fewer points are upsampled by duplicating randomly selected points.
This processed set of points is used by two dedicated LiDAR encoders.
A point-based encoder operates directly on the $N$ three-dimensional points, while a parallel depth-based encoder projects the same points into an image plane to generate a single-channel depth map at the model's input resolution $H'\times W'$. This is achieved by scaling the camera intrinsics through multiplication with a scale matrix $R'$.
\begin{equation}
R' = \operatorname{diag}\!\big(\frac{W'}{W},\, \frac{H'}{H},\, 1\big), \quad
\mathbf{K'_{\mathrm{Cam}}} = R'\,\mathbf{K_{\mathrm{Cam}}}
\label{eq:SDP}
\end{equation}

We call this projection \textit{scaled depth projection} (SDP).
It is essential to our architecture and proved impactful in our experiments (\cf \cref{sec:experiments:ablation}).
In comparison to projection with the original intrinsic followed by resizing of the depth map, SDP introduces fewer blurring artifacts, which are harmful for precise feature alignment.

\paragraph{RGB and Event Camera.}
For the RGB image input, we apply per-channel standardization using statistics of the training set by subtracting the channel means and dividing by the corresponding standard deviations. We resize the images to the model’s input resolution ($H'\times W'$) using bilinear interpolation and feed them to the RGB feature encoder.
For events, we accumulate the stream over a temporal window of 50ms to construct two-channel frames with positive and negative polarities. 
After time synchronization with the RGB and LiDAR clocks, we slice events in a window centered on each RGB timestamp. 
As stated in our prerequisite, the event and RGB cameras are assumed to be extrinsically calibrated.
We apply the known extrinsics (\cf \cref{sec:method:problem}) to transform the event data into the RGB camera coordinate system, ensuring alignment between the event frame and the RGB image.
Finally, we resize the two-channel event frame to the model’s input resolution using bilinear interpolation.

\subsubsection{Encoding} \label{sec:method:architecture:encoding}

\paragraph{Shared LiDAR Branch.}
The shared LiDAR branch extracts complementary geometric information using two parallel encoders, one \textit{point-based} and one \textit{depth-based}, as shown in the green block of \cref{fig:architecture}.
The point-based encoder uses Point-Transformer-V3~(PTV3)~\cite{ptv3}, which processes unordered 3D points by serializing them through space-filling curves to enable efficient local attention and capture fine-grained geometric structure while remaining computationally scalable.
The depth-based encoder employs Mobile-Vision-Transformer-V2~(MViTV2)~\cite{mvitv2} to capture spatial context from the projected depth maps.

The per-point features produced by the point-based branch are projected onto an image plane by scaling the camera intrinsics based on \cref{eq:SDP}, ensuring consistency with the resolution of the depth features $H''\times W''$.
We refer to this process as \textit{scaled feature projection} (SFP).
Similar to SDP, SFP is beneficial for the same underlying reasons.

The projected point features and depth features are brought to a common channel dimension and fused through channel-wise concatenation, forming a unified LiDAR embedding that combines detailed 3D structure from point features and dense geometric cues from depth features. 
This fusion improves rotation and translation estimation, as confirmed by our experiments.
The shared LiDAR branch benefits both the \LiRGB and \LiEvent branches and results in improved efficiency with lower inference time, parameter count, and memory usage, compared to having separate LiDAR branches for each modality pair (see \cref{tab:splitvscom}).

\paragraph{RGB and Event Encoders.}
The RGB and event branches extract visual features using two independent encoders based on Mobile-Vision-Transformer-V2 (MViTV2)~\cite{mvitv2}. 
MViTV2 is a hybrid architecture that combines convolutional layers with transformer blocks, enabling efficient modeling of local spatial patterns together with global context. 
The convolutional layers capture fine-grained textures and edges, while the transformer components aggregate information across the image to encode long-range dependencies. 
The RGB and event encoders process their respective preprocessed inputs, each independently learning modality-specific representations that capture appearance and motion cues critical for accurate geometric calibration.

\subsubsection{Pair-wise Cost Volumes}
After feature extraction, two parallel modality-specific paths (\LiRGB and \LiEvent) construct correlation cost volumes to measure local cross-modal similarity, following PWC-Net \cite{pwc} and LCCNet \cite{LCCNet}.
Let $\mathbf{F}^{\mathrm{Li}}, \mathbf{F}^{Cam \in \{\mathrm{RGB}, \mathrm{Ev}\}} \in \mathbb{R}^{H''\times W'' \times C}$ be the two feature maps of the shared LiDAR branch and one of the camera branches.
For a pixel $\mathbf{p}=(x,y)$ and a displacement $(\Delta x,\Delta y) \in\{-d,\ldots,d\}$ within a square window of radius $d$, the matching cost is the channel-normalized inner product as defined in the following equation:
\begin{equation}
\mathcal{C}(y,x,\Delta x,\Delta y)
=\frac{1}{C}\sum_{c=1}^{C}
\mathbf{F}^{\mathrm{Li}}_{c,y,x} \cdot 
\mathbf{F}^{\mathrm{Cam}}_{c,\,y+\Delta y,\,x+\Delta x}
\label{eq:cost_volume}
\end{equation}
The costs over all displacements form a three-dimensional cost volume $\mathcal{C}\in\mathbb{R}^{ H''\times W'' \times M}$. Here, $M$ is the number of offset locations in the square window $M=(2d+1)^2$.
The cost volume computes the correlation between the two feature vectors for every pixel and offset.
We then apply a LeakyReLU activation and refine the volume using a context module before predicting the extrinsics.

\subsubsection{Context Modules}
For the two modality-specific paths (\LiRGB and \LiEvent), the context module has identical topologies and follows the design in LCCNet \cite{LCCNet}.
Each consists of five convolutional layers, and after each layer, the output features are concatenated with the running tensor, as in DenseNet \cite{huang2017densely}.
This enhances the channel capacity and preserves early representation, resulting in richer pair-specific latent features.
Concatenating features -- rather than overwriting them -- gradually expands the channel dimension while retaining both low- and high-level information.
The resulting expressive representation is then passed to the fully connected prediction heads.

\subsubsection{Prediction Heads}
Similar to the pair-wise cost volume and pair-wise context modules, we use two identical calibration heads, one for the \LiRGB path and one for the \LiEvent path.
For each case, the resulting pair-specific latent features from the dense convolutional layers are flattened into a one-dimensional vector and passed through a shared fully connected layer. 
The resulting representation is then split into two fully connected prediction heads: one for translation and one for rotation. 
The translation head uses two fully connected layers with a LeakyReLU in between to regress the three-dimensional translation vector.
The rotation head follows a similar design with two fully connected layers and a LeakyReLU in between, but its output is normalized to enforce a valid quaternion representation.
This split prediction head design enables the network to learn translation and rotation simultaneously.

\subsection{Iterative Refinement} \label{sec:method:stages}

Inspired by \cite{LCCNet}, we train multiple independent models, each specialized for different random translation and rotation errors within a specific range, starting from the largest to the smallest error range (\cf \cref{sec:exp:details:perturbation}). 
During evaluation, each stage predicts a transformation that partially corrects the miscalibration, and the partially corrected result serves as input to the next stage. 
Let $k=1,\dots,S$ index the stages, let $\hat{\mathbf T}^{(0)}$ be the initial estimate, representing the identity transformation, \ie, the initial miscalibrated pose is not corrected, and let $\Delta\hat{\mathbf T}^{(k)}$ be the residual predicted by stage $k$. We update via:
\begin{equation}
\hat{\mathbf T}^{v,(k)} = \Delta\hat{\mathbf T}^{v,(k)}\,\hat{\mathbf T}^{v,(k-1)}, \qquad k=1,\dots,S
\label{eq:iter-update}
\end{equation}

\subsection{Loss Functions} \label{sec:method:loss}
The network is trained with three losses per modality pair: a translation loss, a rotation loss, and a point cloud distance loss. 
The translation and rotation loss directly supervise the predicted pose parameters, while the point-cloud distance loss enforces geometric alignment between the LiDAR points transformed by the predicted pose and the same points transformed by the ground truth pose.  

\paragraph{Translation loss.}  
We use a Smooth L1 loss on the predicted translation vector $\hat{\mathbf t}^{v} \in \mathbb{R}^3$ and the ground truth translation $\mathbf t^{v} \in \mathbb{R}^3$ over a batch of size $B$:  
\begin{equation}
\mathcal{L}^v_{\text{trans}} = \frac{1}{B} \sum_{i=1}^{B} 
\|\hat{\mathbf t}^{v}_i - \mathbf t^{v}_i\|_{1}^{\text{smooth}}
\label{eq:loss_tran}
\end{equation}

\paragraph{Rotation loss.}  
Rotations are supervised using the angular distance $\theta(\cdot,\cdot)$ between predicted unit quaternion $\hat{\mathbf{q}}^v$ and ground truth unit quaternion $\mathbf{q}^v$:  
\begin{equation}
\mathcal{L}^v_{\text{rot}} = \frac{1}{B} \sum_{i=1}^{B} 
\theta(\hat{\mathbf q}^{v}_i, \mathbf q^{v}_i)
\label{eq:loss_rot}
\end{equation}

\paragraph{Point cloud distance loss.}  
To ensure geometric consistency, we measure the alignment of LiDAR points $\mathbf{x} \in P_i$ after applying the predicted and ground truth rigid transformations $\hat{\mathbf{T}}^v$ and $\mathbf{T}^v$:
\begin{equation}
\mathcal{L}^v_{\text{pcd}} = \frac{1}{B}\sum_{i=1}^{B} 
\frac{1}{|P_i|}\sum_{\mathbf{x} \in P_i} 
\|\hat{\mathbf{T}}^{v}(\mathbf{x}) - \mathbf{T}^{v}(\mathbf{x})\|_2
\label{eq:loss_pcd}
\end{equation}

\paragraph{Total loss.}  
The following formulation defines the per-pair loss used for the \LiRGB and \LiEvent calibrations, where $\lambda_t,\lambda_r \ge 0$ control the translation and rotation contributions, and $w \in [0,1]$ balances the point cloud loss:  
\begin{equation}
\mathcal{L}^v = (1-w)(\lambda_t \mathcal{L}^v_{\text{trans}} + \lambda_r \mathcal{L}^v_{\text{rot}}) + w \mathcal{L}^v_{\text{pcd}}
\label{eq:loss_v}
\end{equation}
The total training objective is obtained by summing the losses from both \LiRGB and \LiEvent pairs:
\begin{equation}
\mathcal{L}_{\text{total}} = 
\mathcal{L}^{\mathrm{Li}-\mathrm{RGB}} + 
\mathcal{L}^{\mathrm{Li}-\mathrm{Ev}} 
\label{eq:total_loss}
\end{equation}

\section{Experiments and Results} \label{sec:experiments}

\subsection{Datasets} \label{sec:exp:data}
To ensure a fair comparison with state-of-the-art calibration methods and to achieve our target of handling three sensing modalities, we use the KITTI Odometry dataset~\cite{kitti} and DSEC~\cite{DSEC} datasets.

\subsubsection{KITTI Dataset}
The KITTI dataset \cite{kitti} is widely used for \LiRGB calibration and provides accurate, temporally synchronized data from a Velodyne HDL-64E LiDAR and Point Grey Flea stereo RGB cameras; we use the left RGB camera together with the LiDAR for all experiments.
Since KITTI does not include event data, we generate synthetic events from the RGB frames using V2E \cite{V2E}.
For preprocessing, each LiDAR frame is limited to \(N=20{,}000\) front-facing points, based on the average number of front-facing points across the dataset.

\subsubsection{DSEC Dataset}
The DSEC dataset \cite{DSEC} provides all three sensing modalities required for our work, LiDAR, RGB, and events, captured by a Velodyne VLP-16 LiDAR, a FLIR Blackfly-S RGB stereo setup, and a Prophesee Gen3.1 event stereo setup, respectively. 
We use the left RGB camera, the left event camera, and the LiDAR for all experiments, and synchronize the three modalities using the provided timestamps. 
The LiDAR in DSEC is four times sparser than in KITTI, and the RGB images span diverse scenes with varying lighting and weather conditions, making the dataset more realistic and challenging. 
Each LiDAR frame is limited to \(N=5{,}000\) front-facing points to account for its lower density.

\subsection{Implementation and Training Details} \label{sec:exp:details} 

\subsubsection{Artificial LiDAR Perturbation} \label{sec:exp:details:perturbation}
To simulate miscalibration during training and evaluation, we apply artificial perturbations to the ground truth relative poses.
The network learns to recover the correct extrinsics from these perturbed inputs. 
Two perturbation setups are used for comparability with prior works.
Firstly, we adopt a five-stage range with maximum translation and rotation errors of
$\pm \{%
20^\circ/150\,\text{cm},\allowbreak\ 
10^\circ/100\,\text{cm},\allowbreak\ 
5^\circ/50\,\text{cm},\allowbreak\ 
2^\circ/20\,\text{cm},\allowbreak\ 
1^\circ/10\,\text{cm}\}$, to allow fair comparison with RegNet \cite{RegNet} and LCCNet \cite{LCCNet}.
Secondly, we use a two-stage range of 
$\pm \{%
10^\circ/100\,\text{cm},\allowbreak\ 
1^\circ/10\,\text{cm}\}$, to allow fair comparison with MULiEv~\cite{muliev}.

During training, the first stage is trained from scratch using the largest perturbations within the range, while each subsequent stage is initialized from its predecessor and trained on inputs with progressively smaller perturbations.
At each stage, perturbations are sampled uniformly from the corresponding range.
During evaluation, we apply only one random perturbation from the largest stage, and then pass the input sequentially through all stages.

Importantly, two distinct perturbations are applied to the pair-wise modalities to simulate independent \LiRGB and \LiEvent miscalibrations, ensuring that the two cameras are miscalibrated (\ie, perturbed) differently with respect to the LiDAR during both training and testing (\cf independent LiDAR perturbations in \cref{fig:architecture}).
However, this does not resolve our assumption that the RGB and event cameras must be pre-calibrated.

\subsubsection{Hyperparameters} \label{sec:exp:details:hyperparameters}
We train LiREC-Net using the Adam optimizer \cite{adam} with a learning rate of \(3\times10^{-4}\), which decays by a factor of 0.5 at milestones determined through manual experimentation based on validation loss saturation. 
For DSEC, the first stage is trained for 150 epochs, and each subsequent stage is initialized from the previous one and trained for 70 epochs. 
For KITTI, being a smaller dataset compared to DSEC, the first stage is trained for 120 epochs, and each subsequent stage is trained for 50 epochs. 
More details on the learning rate decay and all other hyperparameters are provided in our supplementary material.
The model is trained with a batch size of 64 on four NVIDIA RTXA6000/L40S GPUs.
For a fair comparison of efficiency, all models are evaluated on the same NVIDIA H200 GPU under identical settings.

\subsection{Evaluation Metrics} \label{sec:eval_metrics}
We evaluate calibration accuracy using the mean translation and rotation errors computed over all test samples $N_s$ between the predicted and ground truth transformations. The translation error e\(_t\) is computed as the mean Euclidean distance between the predicted translation vectors $\hat{\mathbf t}^{v}$ and the ground truth translation vectors ${\mathbf t}^{v}$ (see \cref{eq:et}). The rotation error e\(_r\) is defined as the mean angular difference \(\theta\) between the predicted quaternions \(\hat{q}^{v}\) and the ground truth quaternions \(q^{v}\) (see \cref{eq:er}).
\begin{align}
\text{e}_t &= \frac{1}{N_s}\sum_{i=1}^{N_s} \|\hat{\mathbf{t}}^{v}_i - \mathbf{t}^{v}_i\|_2 \label{eq:et}\\
\text{e}_r &= \frac{1}{N_s}\sum_{i=1}^{N_s} \theta(\hat{q}^{v}_{i}, q^{v}_{i}) \label{eq:er}
\end{align}




\subsection{Comparison to State-of-the-Art} \label{sec:exp:sota}
We evaluate LiREC-Net against several state-of-the-art calibration methods on the KITTI and DSEC datasets, as summarized in \cref{tab:sota_kitti,tab:sota_dsec}.

On KITTI, LiREC-Net achieves competitive calibration accuracy for \LiRGB, outperforming CalibNet~\cite{CalibNet} and RegNet~\cite{RegNet}.
We surpass LCCNet~\cite{LCCNet} in terms of rotation, and achieve a marginally higher translation error with a difference of 21mm.
PseudoCal~\cite{pseudocal} shows a translation difference of 62mm and a rotation difference of $0.06^\circ$ (see \cref{tab:sota_kitti}).
We furthermore outperform LCCRAFT~\cite{LCCRAFT} and present the comparison in the supplementary, since LCCRAFT uses different evaluation metrics.
Considering both translation and rotation metrics, LiREC-Net achieves competitive overall performance with only marginal differences compared to existing methods.
Notably, LiREC-Net achieves 1.80cm / $0.11^\circ$ on \LiRGB and 1.82cm / $0.12^\circ$ on \LiEvent, highlighting its ability to calibrate multiple modality pairs.
Furthermore, LiREC-Net is the first to address \LiEvent calibration on KITTI using synthetic events and therefore establishes the first baseline in this category.

\begin{table}[t]
\centering
\caption{Calibration results on the KITTI dataset when LiREC-Net is trained and evaluated with maximum perturbation of $\pm (20^\circ / 150\,\text{cm})$.}
\label{tab:sota_kitti}
\resizebox{\columnwidth}{!}{
\begin{tabular}{lcc}
\toprule
\multirow{2}{*}{\textbf{Network}} & \textbf{\LiRGB} & \textbf{\LiEvent} \\
&  $e_t$ [cm] / $e_r$ [$^\circ$]\phantom{0} & $e_t$ [cm] / $e_r$ [$^\circ$]\phantom{0} \\
\midrule
RegNet \cite{RegNet}     & 6.00 / 0.28 & --- / --- \\
CalibNet \cite{CalibNet} & 4.34 / 0.41 & --- / --- \\
LCCNet \cite{LCCNet}     & 1.59 / 0.16 & --- / --- \\
PseudoCal \cite{pseudocal} & \textbf{1.18} / \textbf{0.05} & --- / --- \\
LiREC-Net (Ours) & 1.80 / 0.11 & \textbf{1.82} / \textbf{0.12} \\
\bottomrule
\end{tabular}
}
\end{table}

\begin{table}[t]
\centering
\caption{Calibration results on the DSEC dataset. [*] Training/evaluation with 5 stages at $\pm (20^\circ / 150\,\text{cm})$, [\textsuperscript{\(\triangle\)}] with 2 stages at $\pm (10^\circ / 100\,\text{cm})$.}
\label{tab:sota_dsec}
\resizebox{\columnwidth}{!}{
\begin{tabular}{lcc}
\toprule
\multirow{2}{*}{\textbf{Network}} & \textbf{\LiRGB} & \textbf{\LiEvent} \\
&  $e_t$ [cm] / $e_r$ [$^\circ$]\phantom{0} & $e_t$ [cm] / $e_r$ [$^\circ$]\phantom{0} \\
\midrule
MULiEv\textsuperscript{\(\triangle\)} \cite{muliev}         & --- / ---   & \textbf{0.81} / 0.10 \\
LiREC-Net\textsuperscript{\(\triangle\)} (Ours) & 2.62 / 0.30 & 2.05 / 0.25 \\
LiREC-Net\textsuperscript{*} (Ours) & \textbf{2.51} / \textbf{0.14} & 1.18 / \textbf{0.07} \\
\bottomrule
\end{tabular}
}
\end{table}

On DSEC, where event data is more prevalent and captured with a real sensor, LiREC-Net achieves strong calibration accuracy across both \LiRGB and \LiEvent pairs (see \cref{tab:sota_dsec}).
For \LiRGB, no prior work reports results on this dataset. 
Thus, our method provides the first benchmark with a translation and rotation error of 2.51cm and $0.14^\circ$, establishing a new baseline for this modality pair.
For \LiEvent, while MULiEv \cite{muliev} achieves a slightly lower translation error (0.81cm \vs 1.18cm), LiREC-Net attains superior rotation accuracy ($0.07^\circ$ \vs $0.10^\circ$) and performs competitively overall, while additionally handling multiple modalities within a single framework.
These results confirm LiREC-Net as the first unified solution for \LiRGB and \LiEvent calibration on DSEC.
\begin{figure}[t]
\centering

\begin{subfigure}{\columnwidth}
    \centering
    \includegraphics[width=\textwidth]{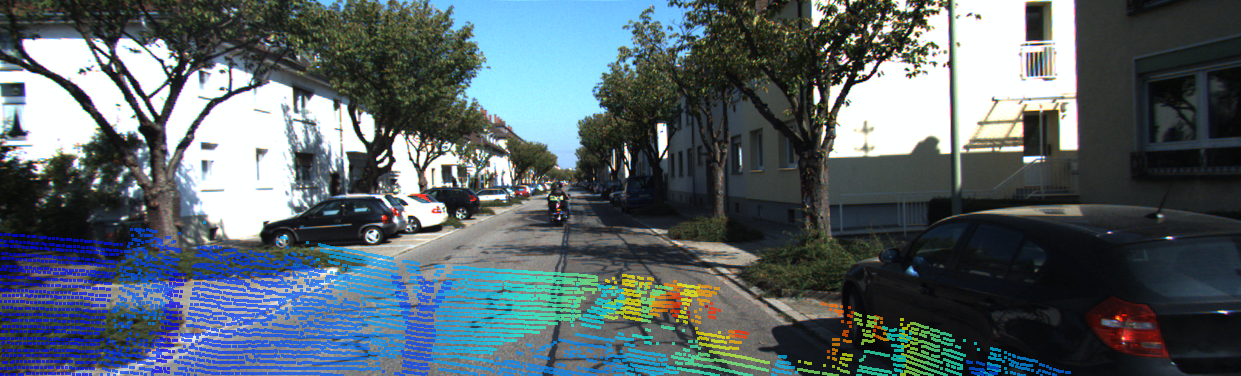}
    \caption{\LiRGB with perturbations within \(\pm(   20^\circ/150\,\text{cm})\).}
\end{subfigure}

\vspace{3pt} 

\begin{subfigure}{\columnwidth}
    \centering
    \includegraphics[width=\textwidth]{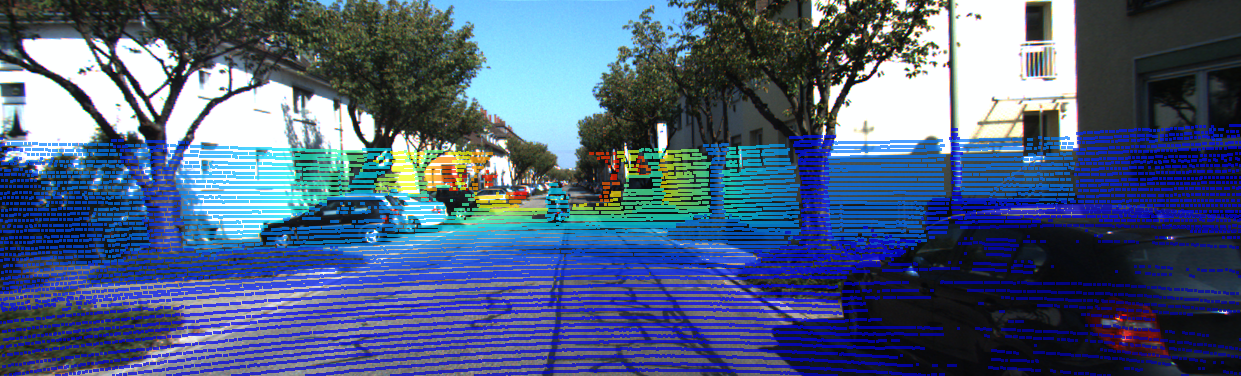}
    \caption{\LiRGB calibration using LiREC-Net's predicted transformation.}
\end{subfigure}

\vspace{3pt} 

\begin{subfigure}{\columnwidth}
    \centering
    \includegraphics[width=\textwidth]{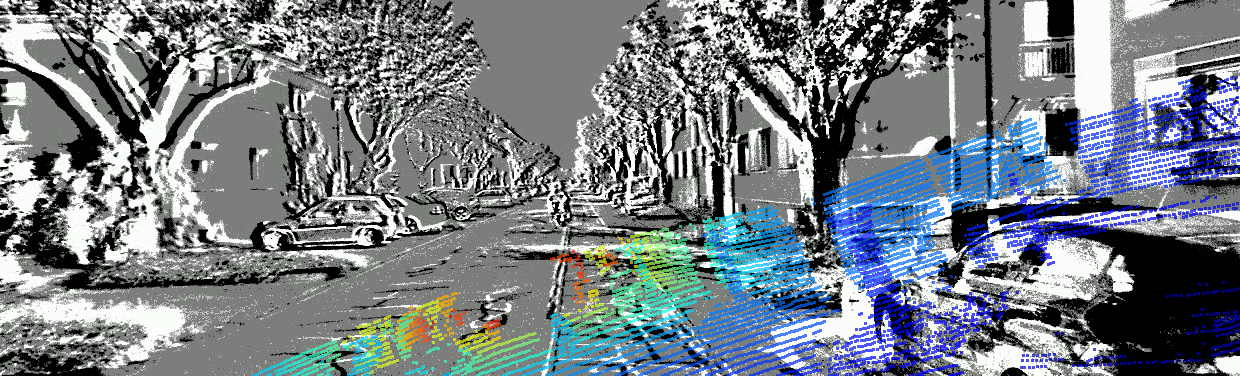}
    \caption{\LiEvent with perturbations within \(\pm (  20^\circ/150\,\text{cm})\).}
\end{subfigure}

\vspace{3pt} 

\begin{subfigure}{\columnwidth}
    \centering
    \includegraphics[width=\textwidth]{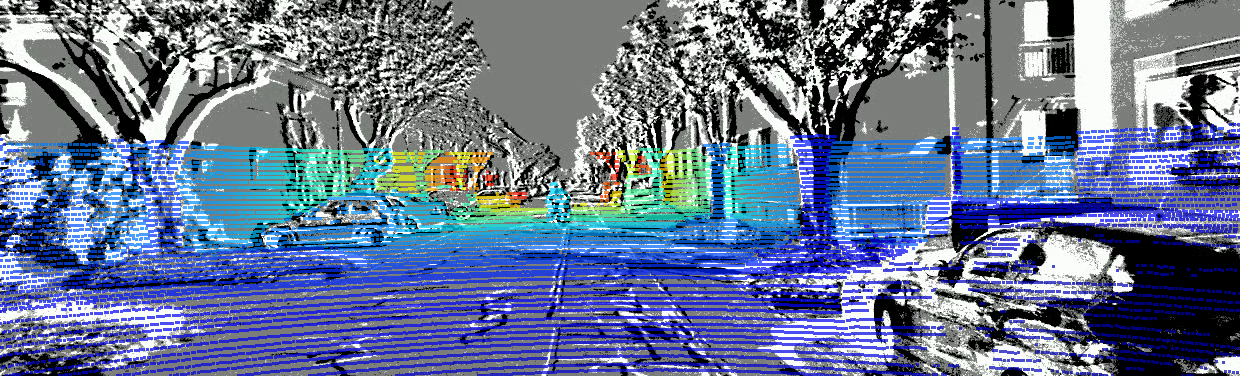}
    \caption{\LiEvent calibration using LiREC-Net's predicted transformation.}
\end{subfigure}

\caption{Qualitative results on KITTI \cite{kitti} for both \LiRGB and \LiEvent pairs.}
\label{fig:kitti-pred}
\end{figure}

\begin{figure}[t]
\centering
\captionsetup[subfigure]{justification=centering}

\begin{subfigure}[t]{0.23\textwidth}
  \centering
  \includegraphics[width=\linewidth]{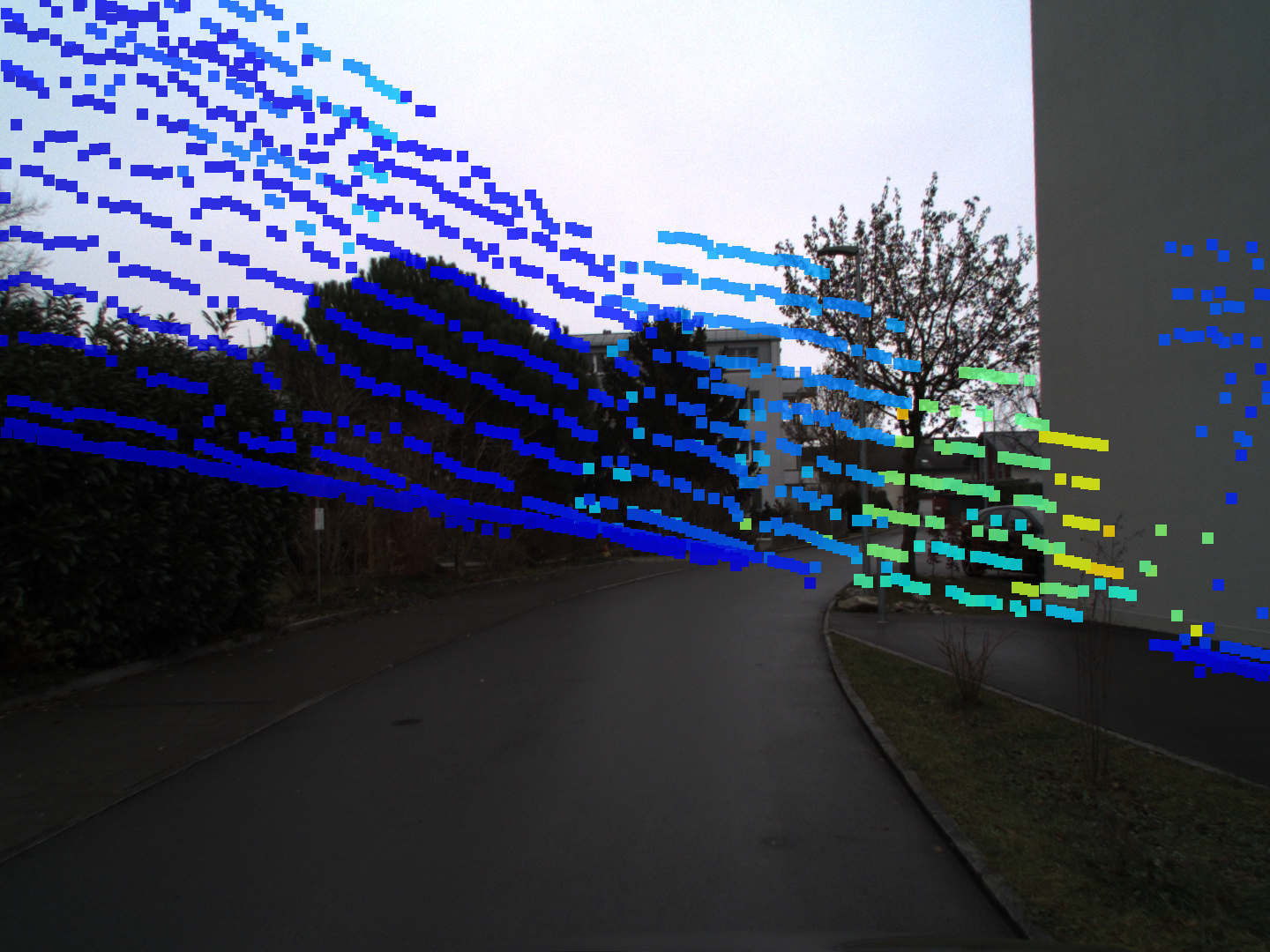}
  \caption{\LiRGB with perturbations within \(\pm(   20^\circ/150\,\text{cm})\).}
\end{subfigure}\hfill
\begin{subfigure}[t]{0.23\textwidth}
  \centering
  \includegraphics[width=\linewidth]{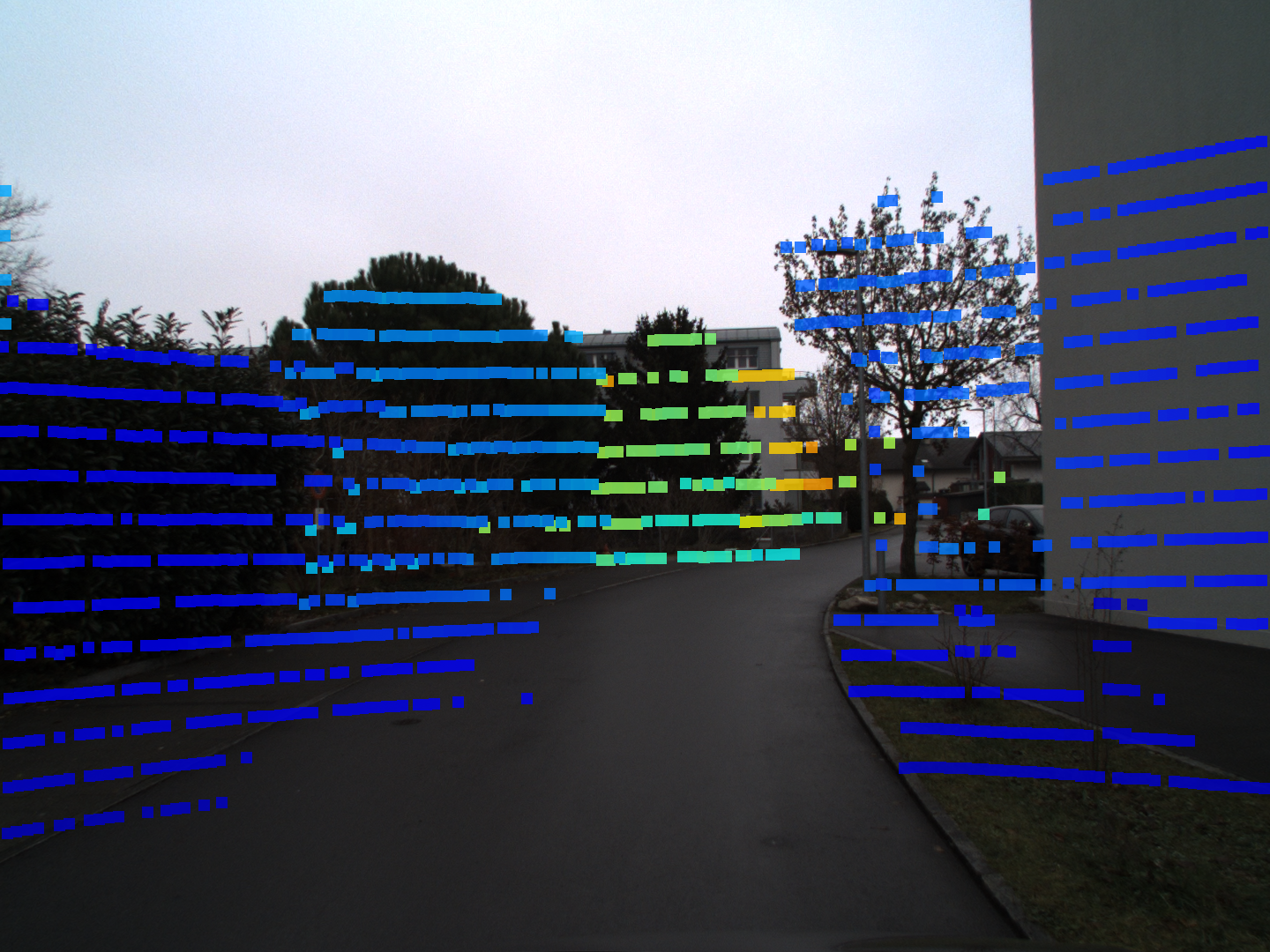}
  \caption{\LiRGB calibration using LiREC-Net's predicted transformation.}
\end{subfigure}

\vspace{3pt} 

\begin{subfigure}[t]{0.23\textwidth}
  \centering
  \includegraphics[width=\linewidth]{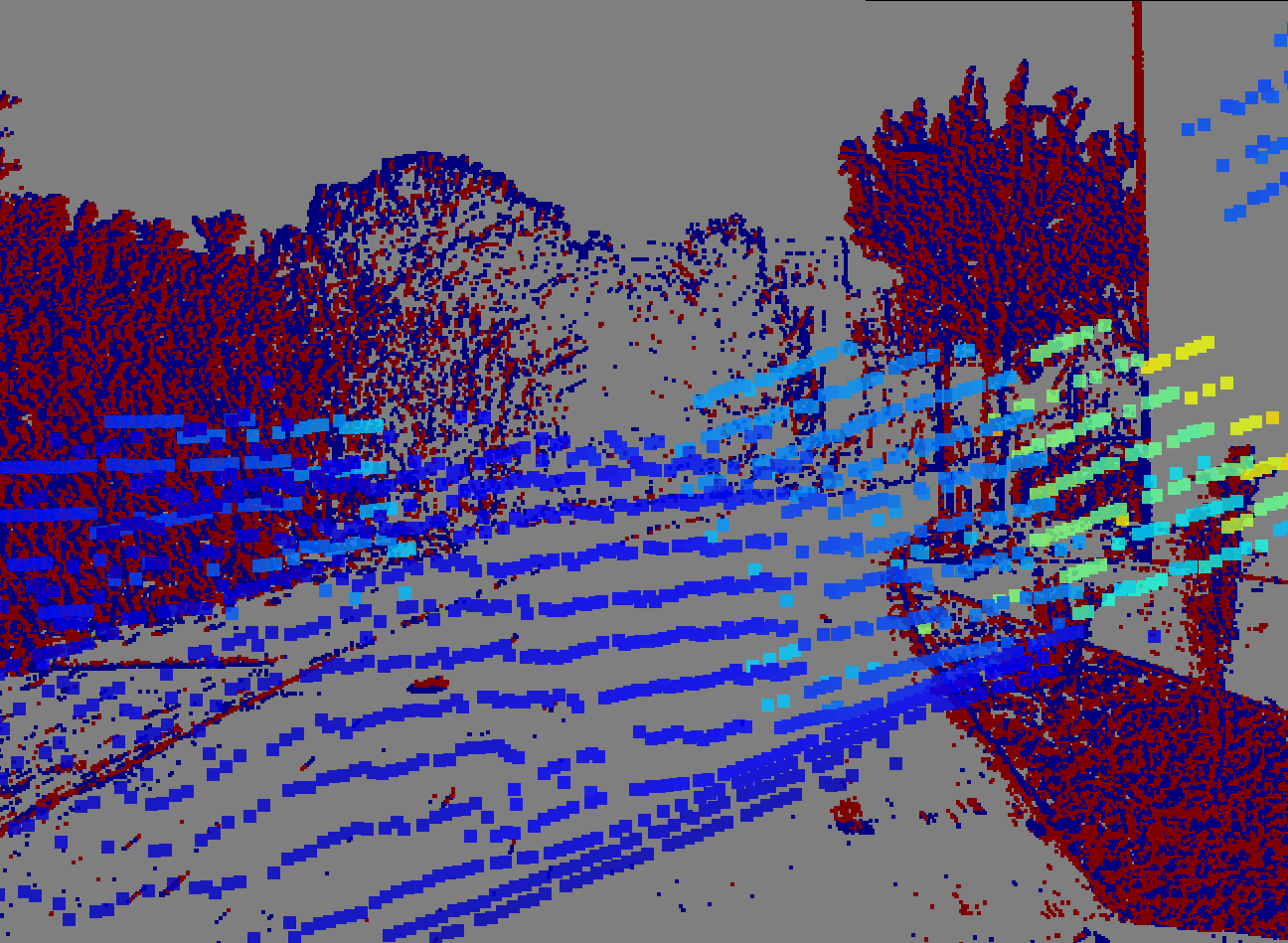}
  \caption{\LiEvent with perturbations within \(\pm(   20^\circ/150\,\text{cm})\).}
\end{subfigure}\hfill
\begin{subfigure}[t]{0.23\textwidth}
  \centering
  \includegraphics[width=\linewidth]{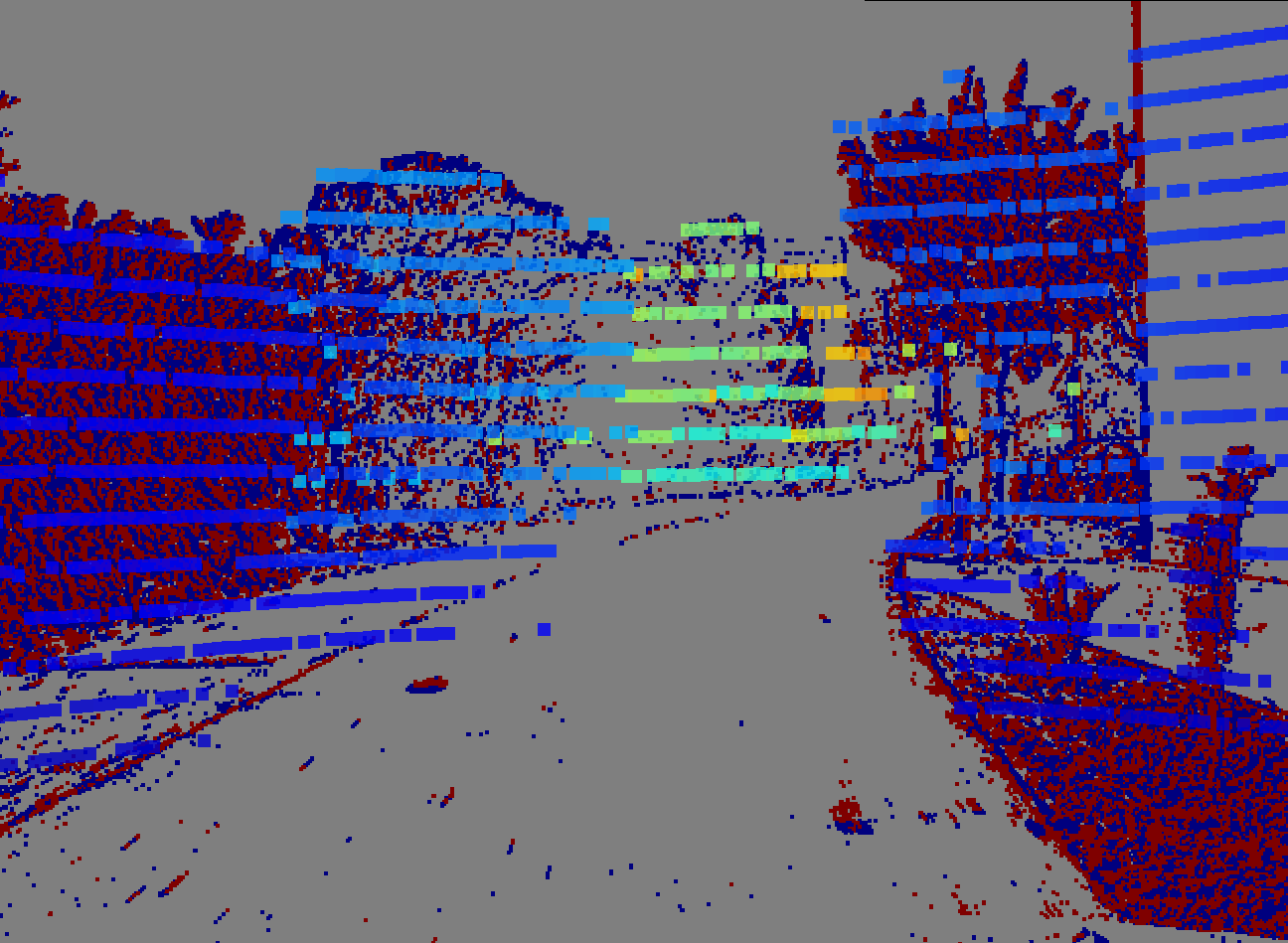}
  \caption{\LiEvent calibration using LiREC-Net's predicted transformation.}
\end{subfigure}

\caption{Qualitative results on DSEC \cite{DSEC} for both \LiRGB and \LiEvent pairs.}
\label{fig:dsec-pred}
\end{figure}

\Cref{fig:kitti-pred,fig:dsec-pred} illustrate qualitative results on the KITTI and DSEC datasets for both \LiRGB and \LiEvent pairs. 
The visualizations each show the projected LiDAR point cloud on top of the respective camera frame before and after calibration with LiREC-Net.
Across both datasets and modalities, LiREC-Net effectively restores spatial alignment between LiDAR projections and the corresponding frames, demonstrating accurate and robust cross-modal calibration.
More qualitative results are provided in our supplementary material.

\subsection{Bi-modal vs.~Tri-modal Calibration}
We analyze the impact of jointly training LiREC-Net on both \LiRGB and \LiEvent (\ie, \textit{tri-modal}), compared to training separate networks for each modality pair (\ie, \textit{bi-modal}), where the latter refers to independent training and evaluation for each pair (\cf \cref{tab:splitvscom}).
For the bi-modal case, the reported inference time (T), parameter count (P), and GPU memory (M) are summed over the two separate models.
For a fair comparison of efficiency, all models are evaluated using the experimental setup detailed in ~\cref{sec:exp:details:hyperparameters}.
The tri-modal setting demonstrates comparable or superior calibration accuracy with improved efficiency.
This efficiency gain stems from the shared LiDAR branch, which avoids duplicated parameters and allows for more efficient computation without compromising on performance.

\begin{table}[t]
\centering
\caption{Calibration result comparison when LiREC-Net is trained and evaluated in tri-modal and bi-modal settings. [*] Training/evaluation with 5 stages at $\pm (20^\circ / 150\,\text{cm})$, [\textsuperscript{\(\triangle\)}] with 2 stages at $\pm (10^\circ / 100\,\text{cm})$. T is the average inference time, P is the total number of model parameters, and M is the GPU memory used for inference.}
\label{tab:splitvscom}
\renewcommand{\arraystretch}{1.1}
\resizebox{\columnwidth}{!}{
\begin{tabular}{lccccc}
\toprule
\multirow{2}{*}{\textbf{Network}} & \textbf{\LiRGB} & \textbf{\LiEvent} & \textbf{T} & \textbf{P} & \textbf{M} \\
&  $e_t$ [cm] / $e_r$ [$^\circ$]\phantom{0} & $e_t$ [cm] / $e_r$ [$^\circ$]\phantom{0} & [s] & [$\times10^9$] & [GiB] \\
\midrule
\multicolumn{6}{c}{\textit{KITTI} \cite{kitti}} \\
\midrule
Bi-modal\textsuperscript{*}          & 2.08 / \textbf{0.11} & 1.90 / \textbf{0.09} & 0.51 & 1.9 & 14.6 \\ 
Tri-modal\textsuperscript{*}       & \textbf{1.80} / \textbf{0.11} & \textbf{1.82} / 0.12 & \textbf{0.33} & \textbf{1.7} & \textbf{11.1} \\ 
\midrule
\multicolumn{6}{c}{\textit{DSEC} \cite{DSEC}} \\
\midrule
Bi-modal\textsuperscript{\(\triangle\)}   & \textbf{2.55} / \textbf{0.28} & 3.43 / \textbf{0.18} & 0.22 & 0.8 & 9.7 \\ 
Tri-modal\textsuperscript{\(\triangle\)} & 2.62 / 0.30 & \textbf{2.05} / 0.25 & \textbf{0.14} & \textbf{0.7} & \textbf{7.1} \\ 
\midrule
Bi-modal\textsuperscript{*} & \textbf{1.40} / \textbf{0.12} & 1.31 / 0.09 & 0.44 & 1.9 & 14.6\\ 
Tri-modal\textsuperscript{*} & 2.51 / 0.14 & \textbf{1.18} / \textbf{0.07} & \textbf{0.31} & \textbf{1.7} & \textbf{11.1}\\ 
\bottomrule
\end{tabular}
}
\end{table}

\subsection{Ablation Study} \label{sec:experiments:ablation}
We perform ablation studies on the DSEC dataset to analyze the contributions of key architectural components and input processing strategies in LiREC-Net. 
We report both translation and rotation errors for \LiRGB and \LiEvent calibration.

We evaluate the impact of LiDAR feature fusion, as shown in \cref{tab:architecture-ablation}.
When either the point or depth feature stream is removed, performance drops sharply. 
Using only point features increases the \LiRGB translation error from 2.51cm to 14.43cm, and the rotation error from 0.14$^\circ$ to 0.70$^\circ$. 
Similarly, using only depth features yields 2.97cm / 0.70$^\circ$ for \LiRGB, showing that both geometric and depth cues are necessary for reliable calibration. 
\LiEvent exhibits a similar trend, confirming that the fusion of point and depth features provides complementary information for accurate alignment.

The effect of the scaled projections (SDP and SFP) in LiREC-Net is analyzed in the studies presented in \cref{tab:scaled-proj-ablation}.
Removing either the scaled depth projection (SDP) or the scaled feature projection (SFP) degrades performance. 
Without SDP, the \LiEvent error rises to 3.72cm / 0.16$^\circ$, while removing SFP leads to 1.45cm / 0.07$^\circ$ for \LiEvent. 
Eliminating both projections results in a significant decline (3.35cm / 0.30$^\circ$ for \LiEvent), confirming their complementary roles in estimating spatial correspondences. 
A similar performance drop is observed with \LiRGB when scaled projections are removed.

Finally, different backbone encoders for RGB, event, and the depth-based LiDAR features are compared, as shown in \cref{tab:resnet-mvitv2-ablation}.
Replacing the transformer-based MViTV2 \cite{mvitv2} with a ResNet \cite{resnet} degrades performance to 3.04cm / 0.19$^\circ$ for \LiRGB and 2.45cm / 0.16$^\circ$ for \LiEvent, underscoring the benefit of global feature modeling for cross-modal alignment. 
Overall, the full LiREC-Net configuration with fused LiDAR features, scaled projections, and the MViTV2 backbone achieves the best performance of 2.51cm / 0.14$^\circ$ for \LiRGB and 1.18cm / 0.07$^\circ$ for \LiEvent.

\begin{table}[t]
\centering
\caption{Ablation showing the effect of LiDAR's point and depth feature fusion on \LiRGB and \LiEvent calibration, conducted on the DSEC dataset.}
\label{tab:architecture-ablation}
\renewcommand{\arraystretch}{1.1}
\resizebox{\columnwidth}{!}{
\begin{tabular}{cccc}
\toprule
\textbf{Point} & \textbf{Depth} & \textbf{\LiRGB} & \textbf{\LiEvent} \\
\textbf{Features} & \textbf{Features} &  $e_t$ [cm] / $e_r$ [$^\circ$]\phantom{0} & $e_t$ [cm] / $e_r$ [$^\circ$]\phantom{0} \\
\midrule
\cmark &  & 14.43 / 0.70 & 14.05 / 0.64 \\
 & \cmark & 2.97 / 0.70 & 2.16 / 0.60 \\
\cmark & \cmark & \textbf{2.51 / 0.14} & \textbf{1.18 / 0.07} \\
\bottomrule
\end{tabular}
}
\end{table}

\begin{table}[t]
\centering
\caption{Ablation showing the effect of scaled projections (SDP and SFP) on \LiRGB and \LiEvent calibration, conducted on DSEC dataset.}
\label{tab:scaled-proj-ablation}
\renewcommand{\arraystretch}{1.1}
\begin{tabular}{cccc}
\toprule
\multirow{2}{*}{\textbf{SDP}} & \multirow{2}{*}{\textbf{SFP}} & \textbf{\LiRGB} & \textbf{\LiEvent} \\
&  & $e_t$ [cm] / $e_r$ [$^\circ$]\phantom{0} & $e_t$ [cm] / $e_r$ [$^\circ$]\phantom{0} \\
\midrule
 &  & 3.37 / 0.34 & 3.35 / 0.30 \\
\cmark &  & 2.95 / 0.15 & 1.45 / \textbf{0.07} \\
 & \cmark & 4.18 / 0.18 & 3.72 / 0.16 \\
\cmark & \cmark & \textbf{2.51 / 0.14} & \textbf{1.18 / 0.07} \\
\bottomrule
\end{tabular}
\end{table}

\begin{table}[t]
\centering
\caption{Ablation showing the effect of MViTV2 and ResNet on \LiRGB and \LiEvent calibration, conducted on DSEC dataset.}
\label{tab:resnet-mvitv2-ablation}
\renewcommand{\arraystretch}{1.1}
\begin{tabular}{lcc}
\toprule
\multirow{2}{*}{\textbf{Encoder}} & \textbf{\LiRGB} & \textbf{\LiEvent} \\
&  $e_t$ [cm] / $e_r$ [$^\circ$]\phantom{0} & $e_t$ [cm] / $e_r$ [$^\circ$]\phantom{0} \\
\midrule
ResNet \cite{resnet} & 3.04 / 0.19 & 2.45 / 0.16 \\
MViTV2 \cite{mvitv2} & \textbf{2.51 / 0.14} & \textbf{1.18 / 0.07} \\
\bottomrule
\end{tabular}
\end{table}

\section{Limitations and Future Work} \label{sec:limitations}
LiREC-Net currently assumes that the RGB and event cameras are pre-calibrated, \ie, the relative pose $\mathbf{T}^{\mathrm{Ev}\rightarrow\mathrm{RGB}}$ is known.
Though our dual perturbation strategy nullifies this condition \wrt the LiDAR coordinate system, we found it critical that the camera features share a common coordinate system, when using a shared LiDAR representation.
Furthermore, the presented design focuses on three specific sensors (LiDAR, RGB, and Event). As future work, we aim to remove the pre-calibration assumption by jointly estimating $\mathbf{T}^{\mathrm{Ev}\rightarrow\mathrm{RGB}}$ within the pipeline, and to generalize the framework to other sensor combinations beyond the three considered here, potentially incorporating additional modalities, \eg, thermal, radar, or depth.

\section{Conclusion} \label{sec:conclusion}
In conclusion, we propose LiREC-Net, a target-free, learning-based, tri-modal framework that jointly calibrates \LiRGB and \LiEvent pairs in a single architecture.
LiREC-Net fuses point- and depth-based LiDAR features in a shared LiDAR branch, employs modality-specific RGB/event encoders, constructs pair-wise cross-modal cost volumes, and predicts translation and unit-quaternion rotation.
It's design and training strategy enable LiREC-Net to predict highly competitive and efficient extrinsic calibrations for tri-modal sensory systems.

\section*{Acknowledgment} 
This work was partially funded by the Federal Ministry of Research, Technology, and Space Germany under the project COPPER (16IW24009).

{
    \small
    \bibliographystyle{ieeenat_fullname}
    \bibliography{bibliography}
}

\clearpage
\setcounter{page}{1}
\maketitlesupplementary

\setcounter{section}{0}
\renewcommand\thesection{\Alph{section}}

\section{Overview}
Here, we provide the supplementary material to our main paper "\textit{LiREC-Net: A Target-Free and Learning-Based Network for LiDAR, RGB, and Event Calibration}".
We detail the division of data for training and evaluation, list all relevant hyperparameters required for the reproduction of the results, and include more quantitative and qualitative results and comparisons.

\section{Dataset Splits}
\Cref{tab:kitti_split,tab:dsec_split} summarize the training and evaluation splits used for the KITTI \cite{kitti} and DSEC \cite{DSEC} datasets, respectively. Each table lists the sequences included in the training and evaluation sets along with the corresponding number of samples.

\section{Training Hyperparameters}
The training schedule for KITTI and DSEC datasets, detailing the number of epochs and learning rate decay milestones per stage, is provided in \cref{tab:lr_schedule}.
Additionally, \cref{tab:supp:hyperparameters} lists the hyperparameters used for all experiments.

\section{Comparison with LCCRAFT} \label{supp:lccraf}

In addition to the results presented in the main paper, we further compare LiREC-Net to LCCRAFT~\cite{LCCRAFT} using the evaluation metrics defined by LCCRAFT. 
Specifically, we report the per-axis translation errors $(e_X, e_Y, e_Z)$ and the roll, pitch, and yaw rotation errors $(e_R, e_P, e_Y)$ in \cref{tab:lccraft-lirec}.

\section{Detailed Results}
\label{supp:detailed-result}
The detailed per-stage results for all modality--dataset combinations are provided in 
\cref{tab:stagewise_lirgb_dsec}--\cref{tab:stagewise_lievent_kitti}. 
All results are obtained under a maximum perturbation of 
$\pm\{20^\circ/150\,\text{cm}\}$ and reflect the 5-stage iterative calibration setup.
For each stage, we report the per-axis translation errors 
$(e_X, e_Y, e_Z)$ and the roll, pitch, and yaw error components $(e_R, e_P, e_Y)$ (\cf \cref{supp:lccraf}) together with the  mean translation error $e_t$ and the mean rotation error $e_r$. 
The stage-wise results reveal a consistent improvement in both translation and rotation accuracy, highlighting the effectiveness of calibration using LiREC-Net for \LiRGB and \LiEvent on both KITTI and DSEC.

\begin{table}[t]
\centering
\caption{Training and evaluation splits for the KITTI dataset \cite{kitti}.}
\label{tab:kitti_split}
\begin{tabular}{ccc}
\toprule
\textbf{Split} & \textbf{Sequences} & \textbf{No. of samples} \\
\midrule
Training  & 01--21 & 39011 \\
Evaluation & 00 & \phantom{0}4541 \\
\bottomrule
\end{tabular}
\end{table}

\begin{table}[t]
\centering
\caption{Training and evaluation splits for the DSEC dataset \cite{DSEC}.}
\label{tab:dsec_split}
\begin{tabular}{ccc}
\toprule
\textbf{Split} & \textbf{Sequences} & \textbf{No. of samples} \\
\midrule
\multirow{14}{*}{Training} & interlaken\_00\_c--g & \multirow{14}{*}{52727} \\
 & thun\_00\_a & \\
 & zurich\_city\_00\_a--b  & \\
 & zurich\_city\_01\_a--f & \\
 & zurich\_city\_02\_a--e & \\
 & zurich\_city\_03\_a & \\
 & zurich\_city\_04\_a--f & \\
 & zurich\_city\_05\_a--b & \\
 & zurich\_city\_06\_a & \\
 & zurich\_city\_07\_a & \\
 & zurich\_city\_08\_a & \\
 & zurich\_city\_09\_a--e & \\
 & zurich\_city\_10\_a--b & \\
 & zurich\_city\_11\_a--c & \\
\midrule
\multirow{7}{*}{Evaluation} & interlaken\_00\_a--b & \multirow{7}{*}{11204} \\
& interlaken\_01\_a & \\
 & thun\_01\_a--b & \\
& zurich\_city\_12\_a & \\
& zurich\_city\_13\_a--b & \\
& zurich\_city\_14\_a--c & \\
& zurich\_city\_15\_a & \\

\bottomrule
\end{tabular}
\end{table}

\begin{table}[t]
\centering
\caption{Training schedule with epochs and learning rate (LR) milestones per stage.}
\label{tab:lr_schedule}
\resizebox{\columnwidth}{!}{
\begin{tabular}{llcl}
\toprule
\textbf{Dataset} & \textbf{Stage} & \textbf{Epochs} & \textbf{Decay Milestones} \\
\midrule
\multirow{2}{*}{KITTI \cite{kitti}} 
& Stage 1 & 120 & 20, 50, 70, 100 \\
& Other Stages & 50 & 15, 30, 40 \\
\midrule
\multirow{2}{*}{DSEC \cite{DSEC}} 
& Stage 1 & 150 & 100, 110, 120, 130 \\
& Other Stages & 70 & 45, 50, 55, 60 \\
\bottomrule
\end{tabular}
}
\end{table}

\begin{table*}[t]
    \centering
    \caption{Hyperparamters and corresponding values of LiREC-Net.}
    \label{tab:supp:hyperparameters}
    \renewcommand{\arraystretch}{1.1}
    \begin{tabular}{ccc}
    \toprule
    Parameter(s) & Description & Value(s) \\
    \midrule
    \multirow{2}{*}{$H\times W$} & Image size (KITTI) & \phantom{0}$376\times1240$\\
    & Image size (DSEC) & $1080\times1440$\\
    $H'\times W'$ & Model input size & $256\times512$\\
    $H''\times W''$ & Feature size & $16\times32$\\
    $d$ & Correlation radius & 4\\
    \multirow{2}{*}{$\mu_{RGB}$} & Mean value (KITTI) & [0.485, 0.456, 0.406]\\
     & Mean value (DSEC) & [0.265, 0.283, 0.300]\\
    \multirow{2}{*}{$\sigma_{RGB}$} & Standard deviation (KITTI) & [0.229, 0.224, 0.225]\\
     & Standard deviation (DSEC) & [0.245, 0.270, 0.301]\\
    $z_{max}$ & Maximum depth value & 80m\\
    \bottomrule
    \end{tabular}
\end{table*}

\begin{table*}[t]
    \centering
    \caption{Comparison between LCCRAFT \cite{LCCRAFT} and our LiREC-Net on KITTI. Errors are reported per translation axis ($e_X$, $e_Y$, $e_Z$ in cm) and rotation ($e_R$, $e_P$, $e_Y$ in [$^\circ$]).}
    \label{tab:lccraft-lirec}
    \renewcommand{\arraystretch}{1.1}
    \begin{tabular}{lcccccccc}
    \toprule
    \multirow{2}{*}{\textbf{Network}} & \multicolumn{4}{c}{\textbf{Translation}} &  \multicolumn{4}{c}{\textbf{Rotations}} \\ 
    \cmidrule(lr){2-5} \cmidrule(lr){6-9}
    & $e_X$ [cm] & $e_Y$ [cm] & $e_Z$ [cm] & $\frac{e_X, e_Y, e_Z}{3}$ [cm] & $e_R$ [$^\circ$] & $e_P$ [$^\circ$] & $e_Y$ [$^\circ$] & $\frac{e_R, e_P, e_Y}{3}$ [$^\circ$]\\
    \midrule
    LCCRAFT \cite{LCCRAFT} & \textbf{0.821} & 0.771 & 1.631 & 1.074 & \textbf{0.029} & 0.117 & 0.041 & 0.062\\
    LiREC-Net & 1.348 & \textbf{0.523} & \textbf{0.747} & \textbf{0.873} & 0.043& \textbf{0.092} & \textbf{0.030} & \textbf{0.055}\\
    \bottomrule
    \end{tabular}
\end{table*}





\begin{table*}[t]
\centering
\caption{Detailed per-stage results of LIREC-Net.}
\label{tab:stagewise}

\begin{subtable}{\linewidth}
\centering
\caption{\LiRGB on DSEC.}
\label{tab:stagewise_lirgb_dsec}
\begin{tabular}{c cccc cccc}
\toprule
\multirow{2}{*}{\textbf{Stage}} & \multicolumn{4}{c}{\textbf{Translation}} & \multicolumn{4}{c}{\textbf{Rotation}} \\
\cmidrule(lr){2-5} \cmidrule(lr){6-9}
 & $e_X$ [cm] & $e_Y$ [cm] & $e_Z$ [cm] & $e_t$ [cm] & $e_R$ [$^\circ$] & $e_P$ [$^\circ$] & $e_Y$ [$^\circ$] &  $e_r$ [$^\circ$] \\
\midrule
1 & 9.54 & 2.33 & 4.39 & 11.82 & 0.24 & 0.48 & 0.31 & 0.72 \\
2 & 2.93 & 0.93 & 2.22 & 4.23 & 0.10 & 0.46 & 0.13 & 0.54 \\
3 & 1.86 & 0.63 & 1.37 & 2.66 & 0.07 & 0.37 & 0.09 & 0.42 \\
4 & 1.17 & 0.57 & 1.57 & 2.25 & 0.07 & 0.17 & 0.08 & 0.24 \\
5 & 1.05 & 0.70 & 1.95 & 2.51 & 0.06 & 0.07 & 0.07 & 0.14 \\
\bottomrule
\end{tabular}    
\end{subtable}
\vspace{0.5em}

\begin{subtable}{\linewidth}
\centering
\caption{\LiEvent on DSEC.}
\label{tab:stagewise_lievent_dsec}
\begin{tabular}{c cccc cccc}
\toprule
\multirow{2}{*}{\textbf{Stage}} & \multicolumn{4}{c}{\textbf{Translation}} & \multicolumn{4}{c}{\textbf{Rotation}} \\
\cmidrule(lr){2-5} \cmidrule(lr){6-9}
 & $e_X$ [cm] & $e_Y$ [cm] & $e_Z$ [cm] & $e_t$ [cm] & $e_R$ [$^\circ$] & $e_P$ [$^\circ$] & $e_Y$ [$^\circ$] &  $e_r$ [$^\circ$] \\
\midrule
1 & 9.29 & 2.37 & 4.50 & 11.66 & 0.22 & 0.47 & 0.29 & 0.69\\
2 & 2.80 & 0.83 & 2.50 & 4.33 & 0.07 & 0.42 & 0.08 & 0.46\\
3 & 1.61 & 0.53 & 1.38 & 2.44 & 0.05 & 0.33 & 0.06 & 0.36\\
4 & 1.15 & 0.92 & 2.63 & 3.24 & 0.05 & 0.15 & 0.05 & 0.19\\
5 & 0.91 & 0.27 & 0.49 & 1.18 & 0.02 & 0.04 & 0.04 & 0.07\\
\bottomrule
\end{tabular}
\end{subtable}
\vspace{0.5em}

\begin{subtable}{\linewidth}
\centering
\caption{\LiRGB on KITTI.}
\label{tab:stagewise_lirgb_kitti}
\begin{tabular}{c cccc cccc}
\toprule
\multirow{2}{*}{\textbf{Stage}} & \multicolumn{4}{c}{\textbf{Translation}} & \multicolumn{4}{c}{\textbf{Rotation}} \\
\cmidrule(lr){2-5} \cmidrule(lr){6-9}
 & $e_X$ [cm] & $e_Y$ [cm] & $e_Z$ [cm] & $e_t$ [cm] & $e_R$ [$^\circ$] & $e_P$ [$^\circ$] & $e_Y$ [$^\circ$] &  $e_r$ [$^\circ$] \\
\midrule
1 & 5.34 & 3.99 & 5.74 & 10.04 & 0.22 & 0.39 & 0.32 & 0.63\\
2 & 2.89 & 2.01 & 2.32 & 4.76 & 0.10 & 0.17 & 0.11 & 0.26\\
3 & 1.89 & 1.31 & 1.82 & 3.25 & 0.07 & 0.11 & 0.06 & 0.16\\
4 & 1.48 & 0.61 & 1.28 & 2.26 & 0.04 & 0.09 & 0.04 & 0.12\\
5 & 1.35 & 0.52 & 0.75 & 1.80 & 0.04 & 0.09 & 0.03 & 0.11\\
\bottomrule
\end{tabular}
\end{subtable}
\vspace{0.5em}

\begin{subtable}{\linewidth}
\centering
\caption{\LiEvent on KITTI.}
\label{tab:stagewise_lievent_kitti}
\begin{tabular}{c cccc cccc}
\toprule
\multirow{2}{*}{\textbf{Stage}} & \multicolumn{4}{c}{\textbf{Translation}} & \multicolumn{4}{c}{\textbf{Rotation}} \\
\cmidrule(lr){2-5} \cmidrule(lr){6-9}
 & $e_X$ [cm] & $e_Y$ [cm] & $e_Z$ [cm] & $e_t$ [cm] & $e_R$ [$^\circ$] & $e_P$ [$^\circ$] & $e_Y$ [$^\circ$] &  $e_r$ [$^\circ$] \\
\midrule
1 & 4.89 & 3.25 & 3.64 & 7.91 & 0.20 & 0.36 & 0.31 & 0.60\\
2 & 2.75 & 1.79 & 2.88 & 4.91 & 0.07 & 0.17 & 0.07 & 0.22\\
3 & 1.92 & 1.60 & 1.91 & 3.46 & 0.06 & 0.11 & 0.08 & 0.16\\
4 & 1.44 & 0.58 & 1.31 & 2.24 & 0.03 & 0.09 & 0.05 & 0.12\\
5 & 1.30 & 0.51 & 0.88 & 1.82 & 0.04 & 0.09 & 0.04 & 0.12\\
\bottomrule
\end{tabular}
\end{subtable}

\end{table*}

\section{Additional Qualitative Results}
\begin{figure*}[p]
\centering
\begin{subfigure}{0.48\linewidth}
    \centering
    \includegraphics[width=\linewidth]{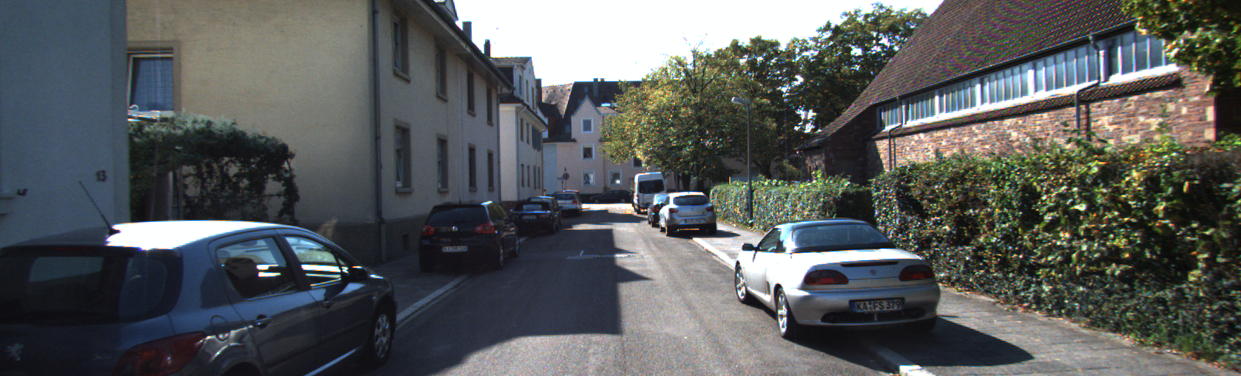}
    \caption{Miscalibrated input [$e_t=173.88\text{cm}, e_r=27.10^\circ$]}
\end{subfigure}
\hspace{1em}
\begin{subfigure}{0.48\linewidth}
    \centering
    \includegraphics[width=\linewidth]{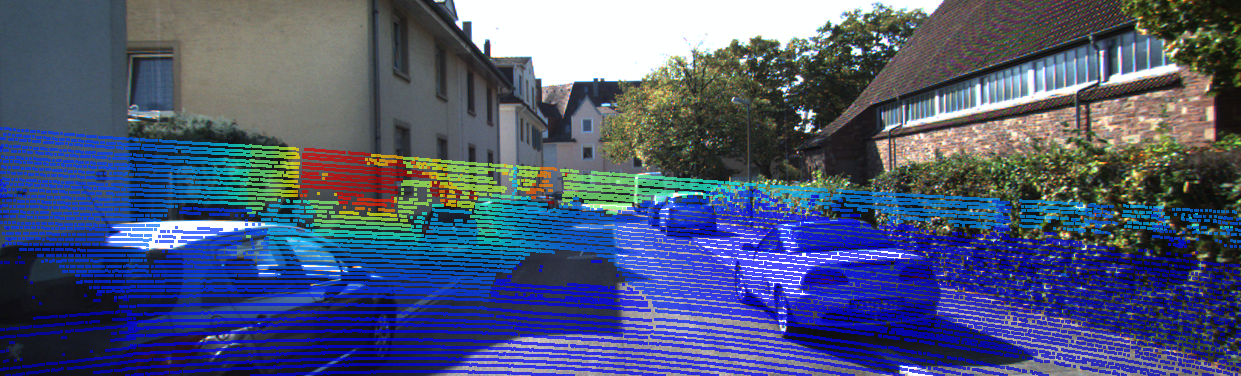}
    \caption{Stage~1 prediction [$e_t=106.92\text{cm}, e_r=18.97^\circ$]}
\end{subfigure}

\vspace{2mm}

\begin{subfigure}{0.48\linewidth}
    \centering
    \includegraphics[width=\linewidth]{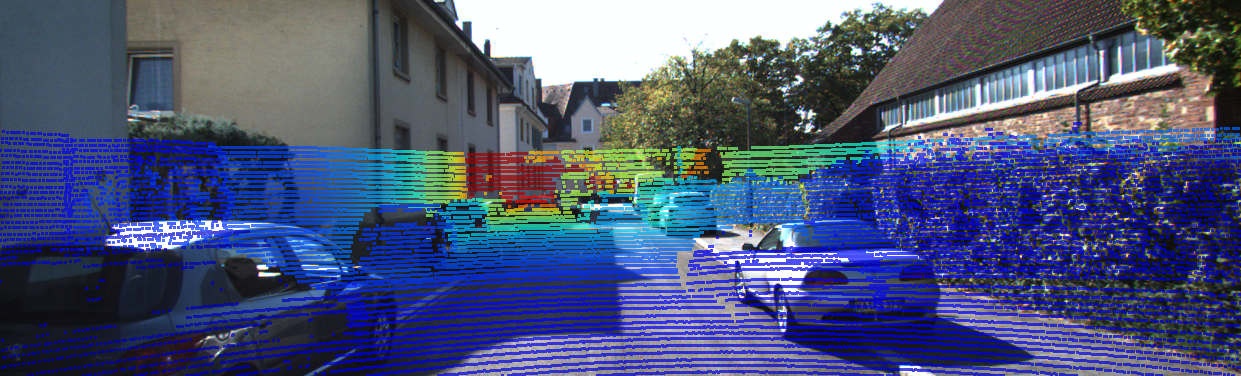}
    \caption{Stage~2 prediction [$e_t=61.72\text{cm}, e_r=6.69^\circ$]}
\end{subfigure}
\hspace{1em}
\begin{subfigure}{0.48\linewidth}
    \centering
    \includegraphics[width=\linewidth]{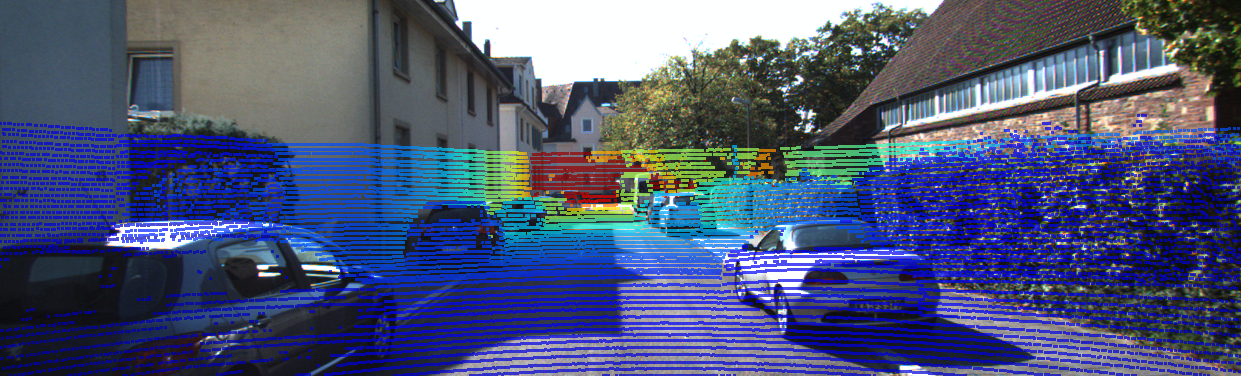}
    \caption{Stage~3 prediction [$e_t=6.07\text{cm}, e_r=1.24^\circ$]}
\end{subfigure}

\vspace{2mm}

\begin{subfigure}{0.48\linewidth}
    \centering
    \includegraphics[width=\linewidth]{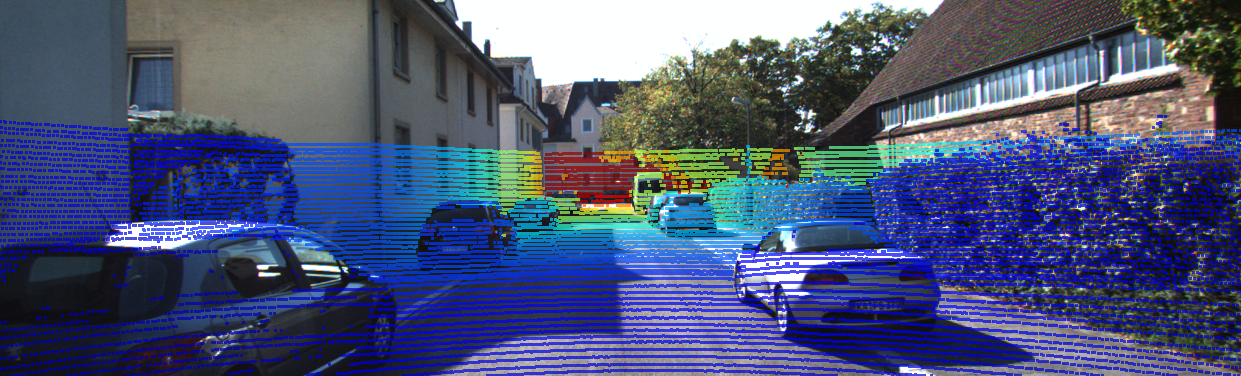}
    \caption{Stage~4 prediction [$e_t=3.60\text{cm}, e_r=0.19^\circ$]}
\end{subfigure}
\hspace{1em}
\begin{subfigure}{0.48\linewidth}
    \centering
    \includegraphics[width=\linewidth]{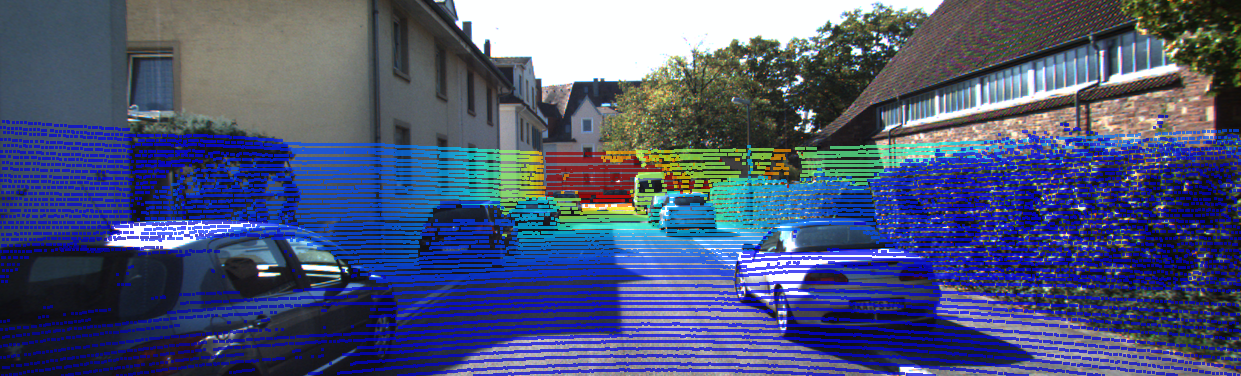}
    \caption{Stage~5 prediction [$e_t=2.81\text{cm}, e_r=0.10^\circ$]}
\end{subfigure}

\vspace{2mm}

\begin{subfigure}{0.48\linewidth}
    \centering
    \includegraphics[width=\linewidth]{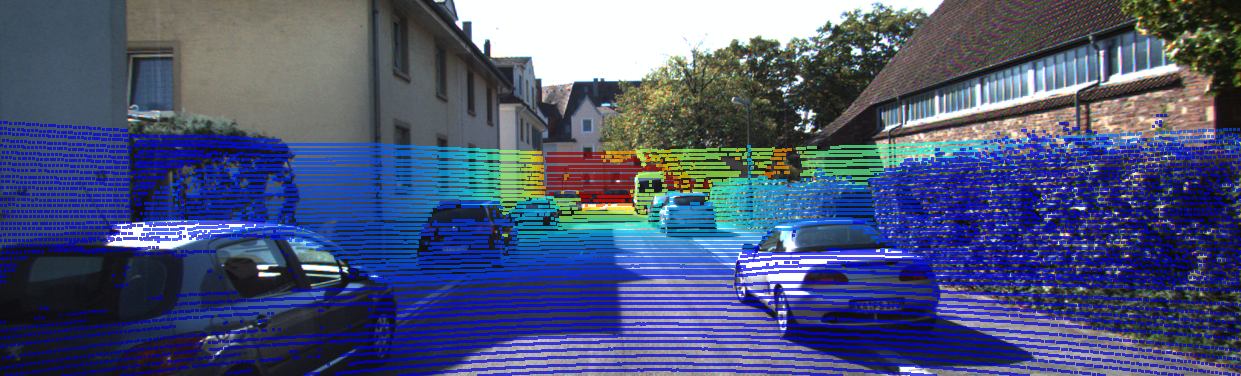}
    \caption{Ground truth}
\end{subfigure}

\caption{Per-stage visualization for \LiRGB calibration on KITTI. 
The exact translation and rotation error is provided for the perturbed input and each stage.
For miscalibrated input, there are no LiDAR points inside the RGB image boundary because of the large perturbation.}
\label{fig:lirgb_kitti_stages}
\end{figure*}

\begin{figure*}[p]
\centering
\begin{subfigure}{0.48\linewidth}
    \centering
    \includegraphics[width=\linewidth]{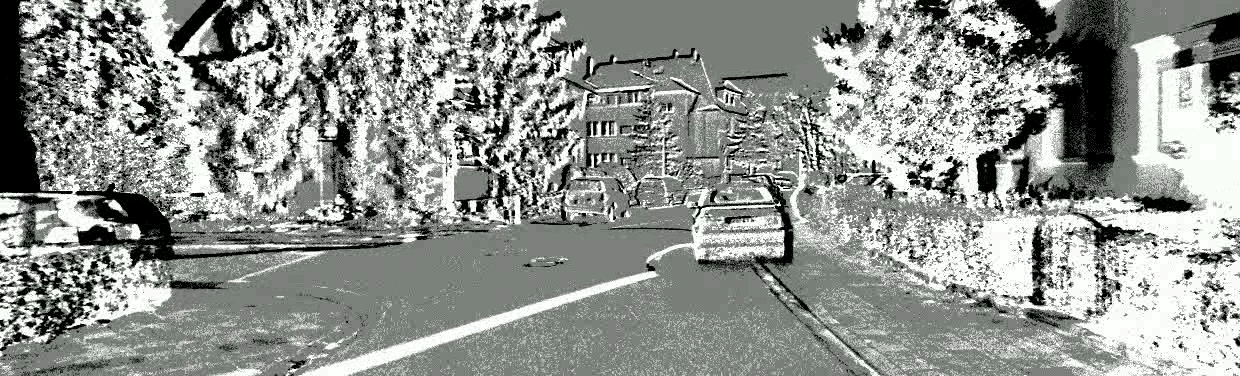}
    \caption{Miscalibrated input [$e_t=177.95\text{cm}, e_r=26.54^\circ$]}
\end{subfigure}
\hspace{1em}
\begin{subfigure}{0.48\linewidth}
    \centering
    \includegraphics[width=\linewidth]{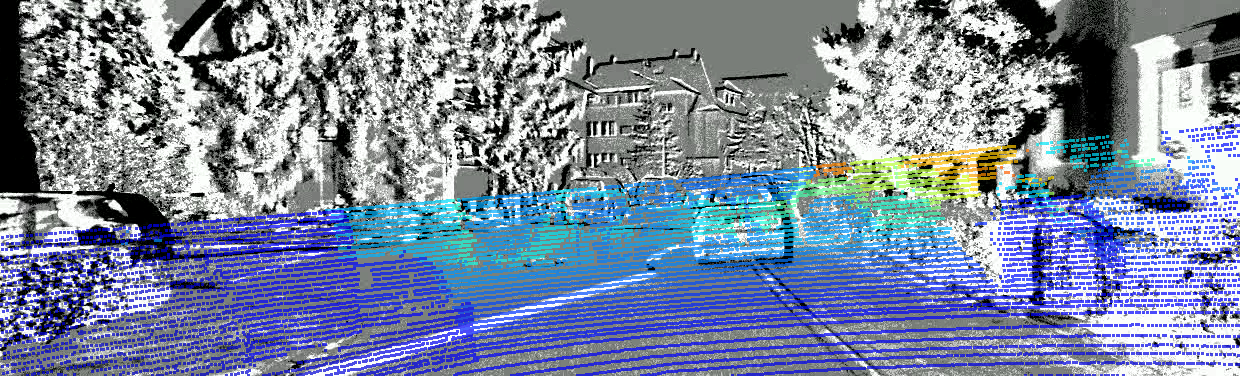}
    \caption{Stage~1 prediction [$e_t=129.84\text{cm}, e_r=16.64^\circ$]}
\end{subfigure}

\vspace{2mm}

\begin{subfigure}{0.48\linewidth}
    \centering
    \includegraphics[width=\linewidth]{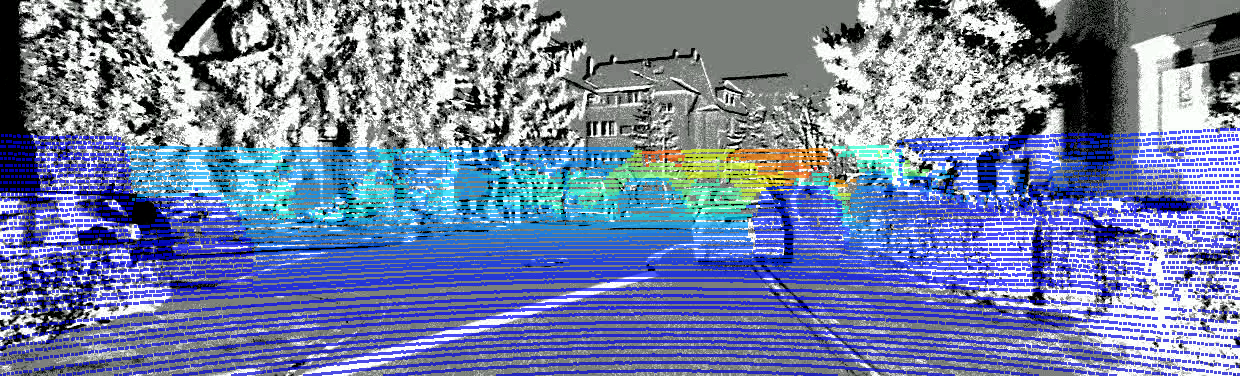}
    \caption{Stage~2 prediction [$e_t=19.93\text{cm}, e_r=3.93^\circ$]}
\end{subfigure}
\hspace{1em}
\begin{subfigure}{0.48\linewidth}
    \centering
    \includegraphics[width=\linewidth]{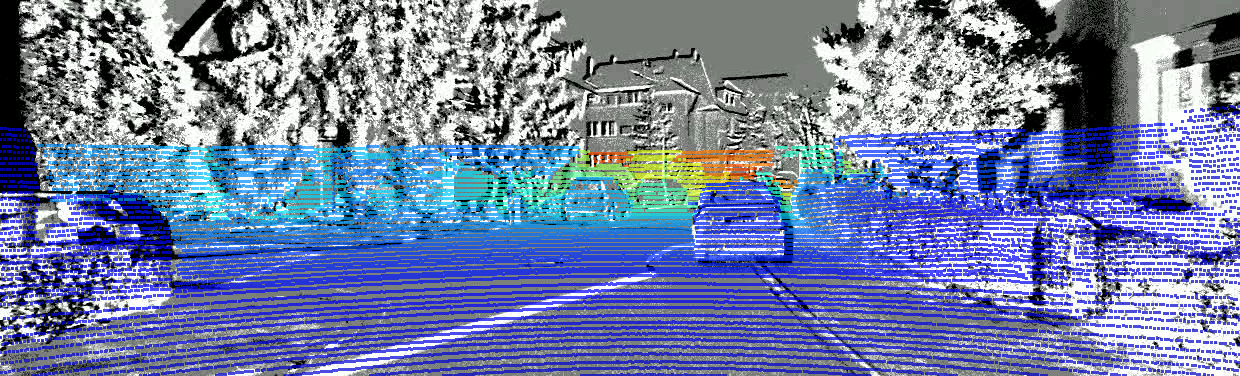}
    \caption{Stage~3 prediction [$e_t=2.62\text{cm}, e_r=0.11^\circ$]}
\end{subfigure}

\vspace{2mm}

\begin{subfigure}{0.48\linewidth}
    \centering
    \includegraphics[width=\linewidth]{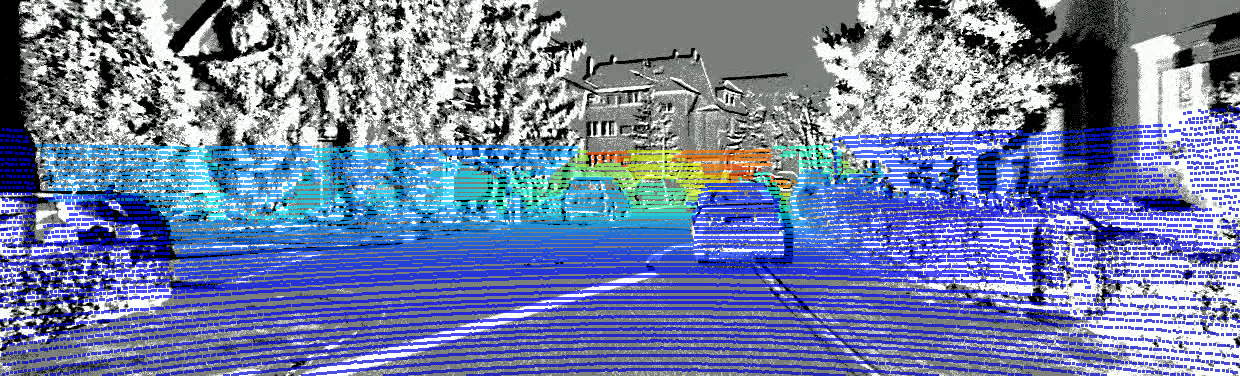}
    \caption{Stage~4 prediction [$e_t=1.90\text{cm}, e_r=0.16^\circ$]}
\end{subfigure}
\hspace{1em}
\begin{subfigure}{0.48\linewidth}
    \centering
    \includegraphics[width=\linewidth]{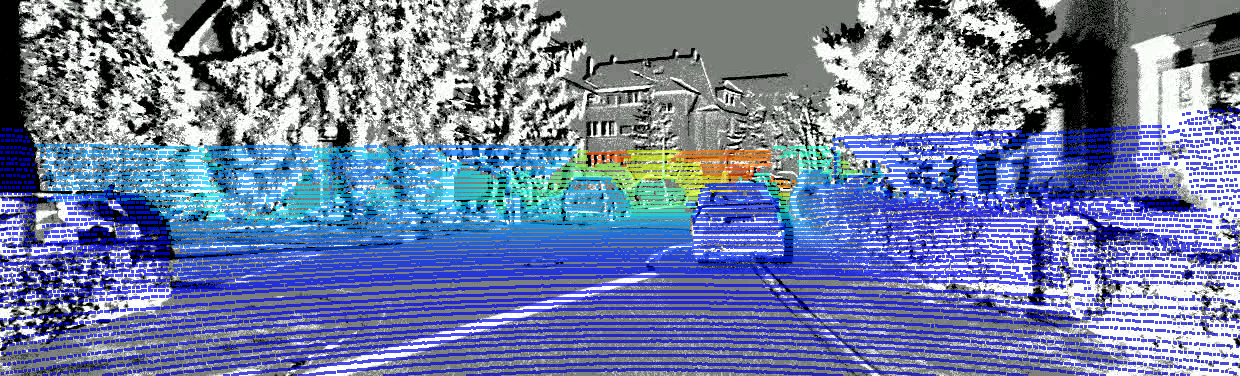}
    \caption{Stage~5 prediction [$e_t=1.62\text{cm}, e_r=0.19^\circ$]}
\end{subfigure}

\vspace{2mm}

\begin{subfigure}{0.48\linewidth}
    \centering
    \includegraphics[width=\linewidth]{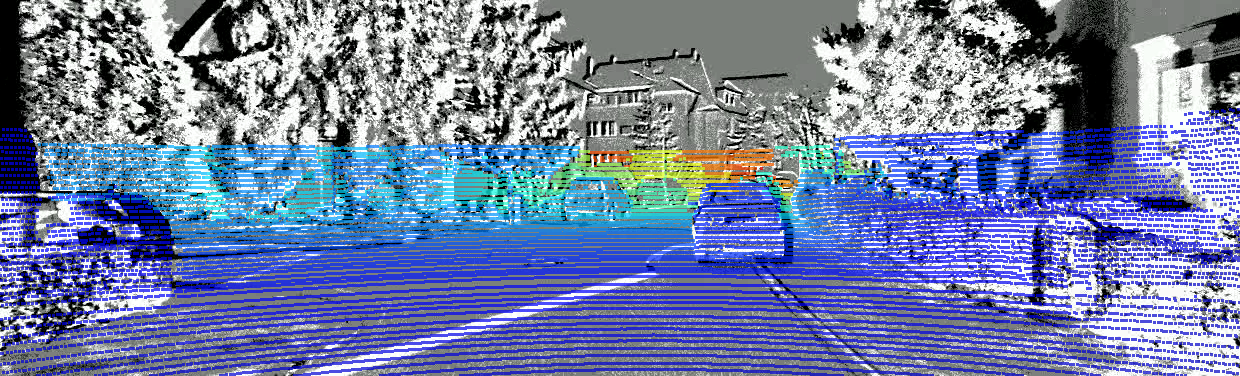}
    \caption{Ground truth}
\end{subfigure}

\caption{Per-stage visualization for \LiEvent calibration on KITTI. 
The exact translation and rotation error is provided for the perturbed input and each stage.
For miscalibrated input, there are no LiDAR points inside the RGB image boundary because of the large perturbation.}
\label{fig:lievent_kitti_stages}
\end{figure*}

\begin{figure*}[p]
\centering

\begin{subfigure}[t]{0.37\linewidth}
    \centering
    \includegraphics[width=\linewidth,trim={0 0 0 2cm},clip]{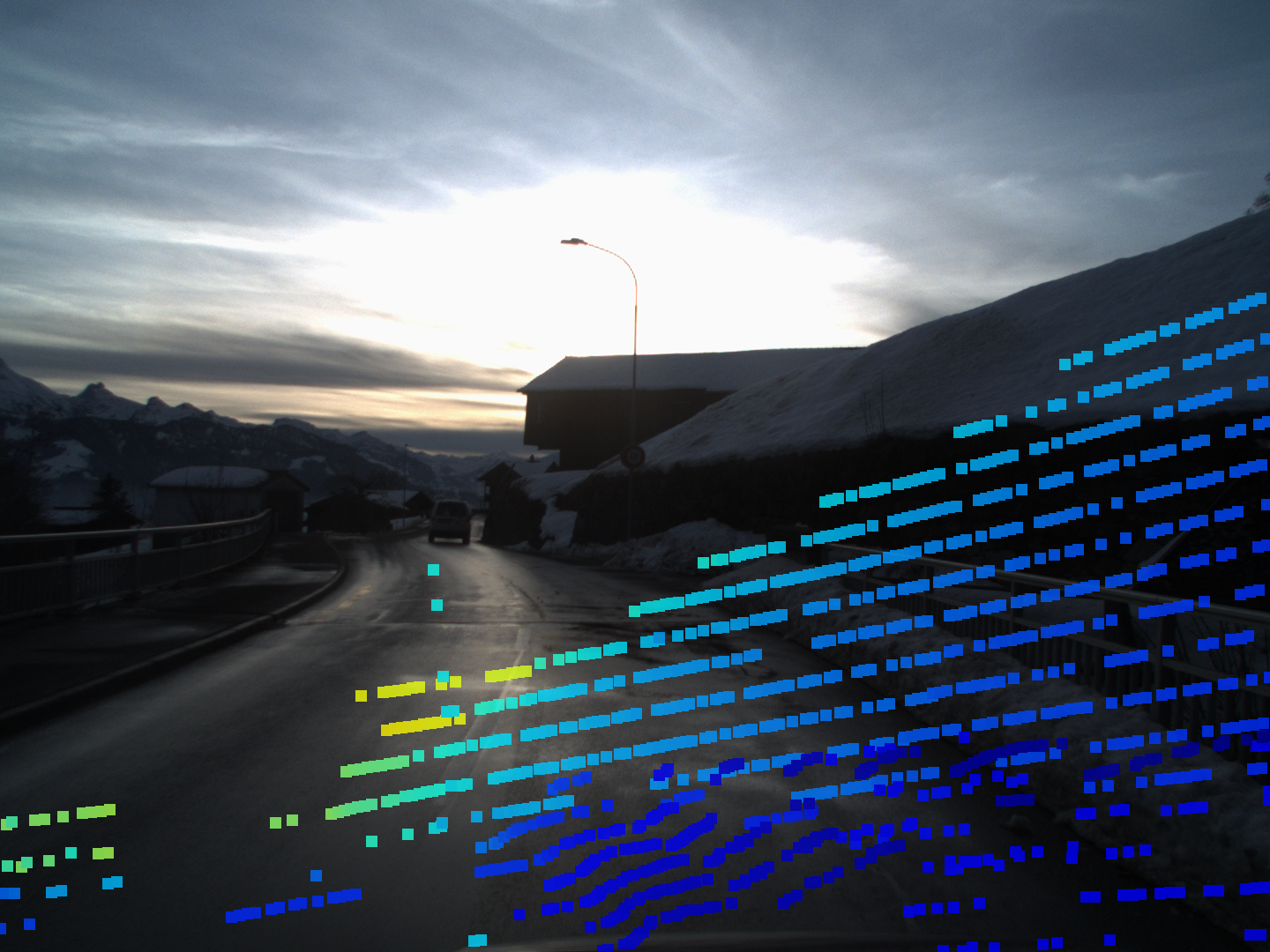}
    \caption{Miscalibrated input [$e_t=207.90\text{cm}, e_r=18.27^\circ$]}
\end{subfigure}
\hspace{1em}
\begin{subfigure}[t]{0.37\linewidth}
    \centering
    \includegraphics[width=\linewidth,trim={0 0 0 2cm},clip]{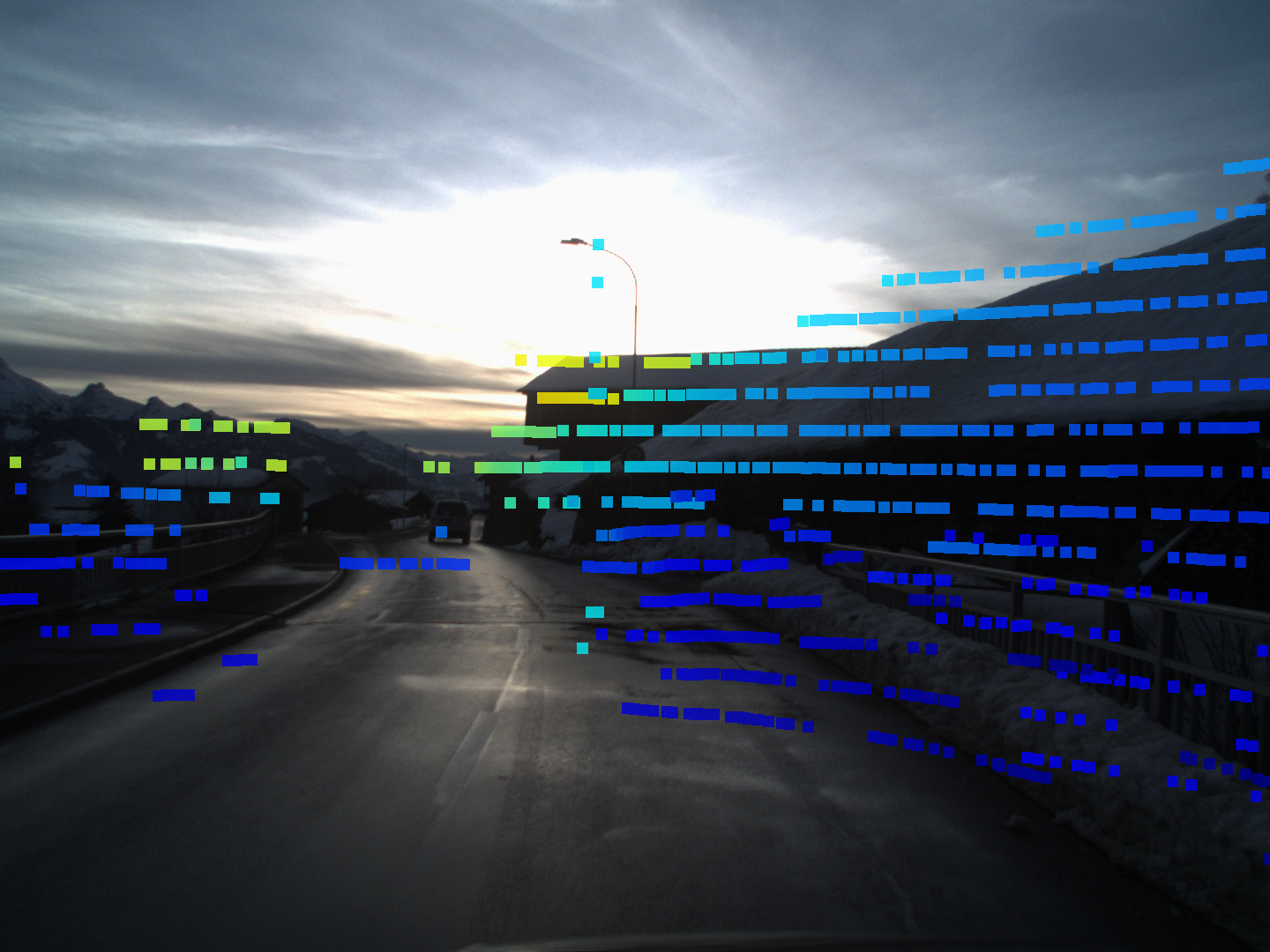}
    \caption{Stage~1 prediction [$e_t=189.45\text{cm}, e_r=3.39^\circ$]}
\end{subfigure}

\vspace{2mm}

\begin{subfigure}[t]{0.37\linewidth}
    \centering
    \includegraphics[width=\linewidth,trim={0 0 0 2cm},clip]{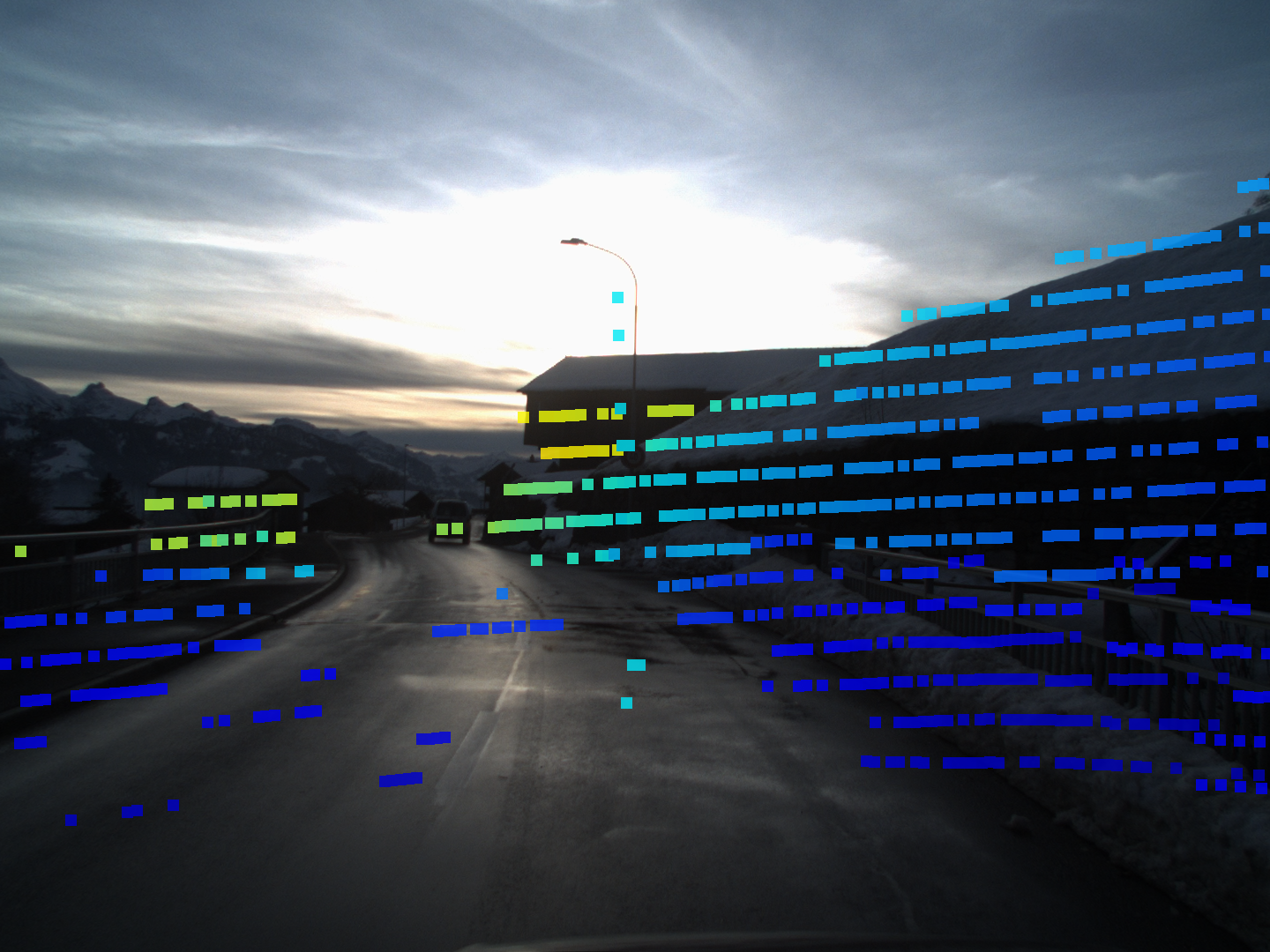}
    \caption{Stage~2 prediction [$e_t=44.22\text{cm}, e_r=1.46^\circ$]}
\end{subfigure}
\hspace{1em}
\begin{subfigure}[t]{0.37\linewidth}
    \centering
    \includegraphics[width=\linewidth,trim={0 0 0 2cm},clip]{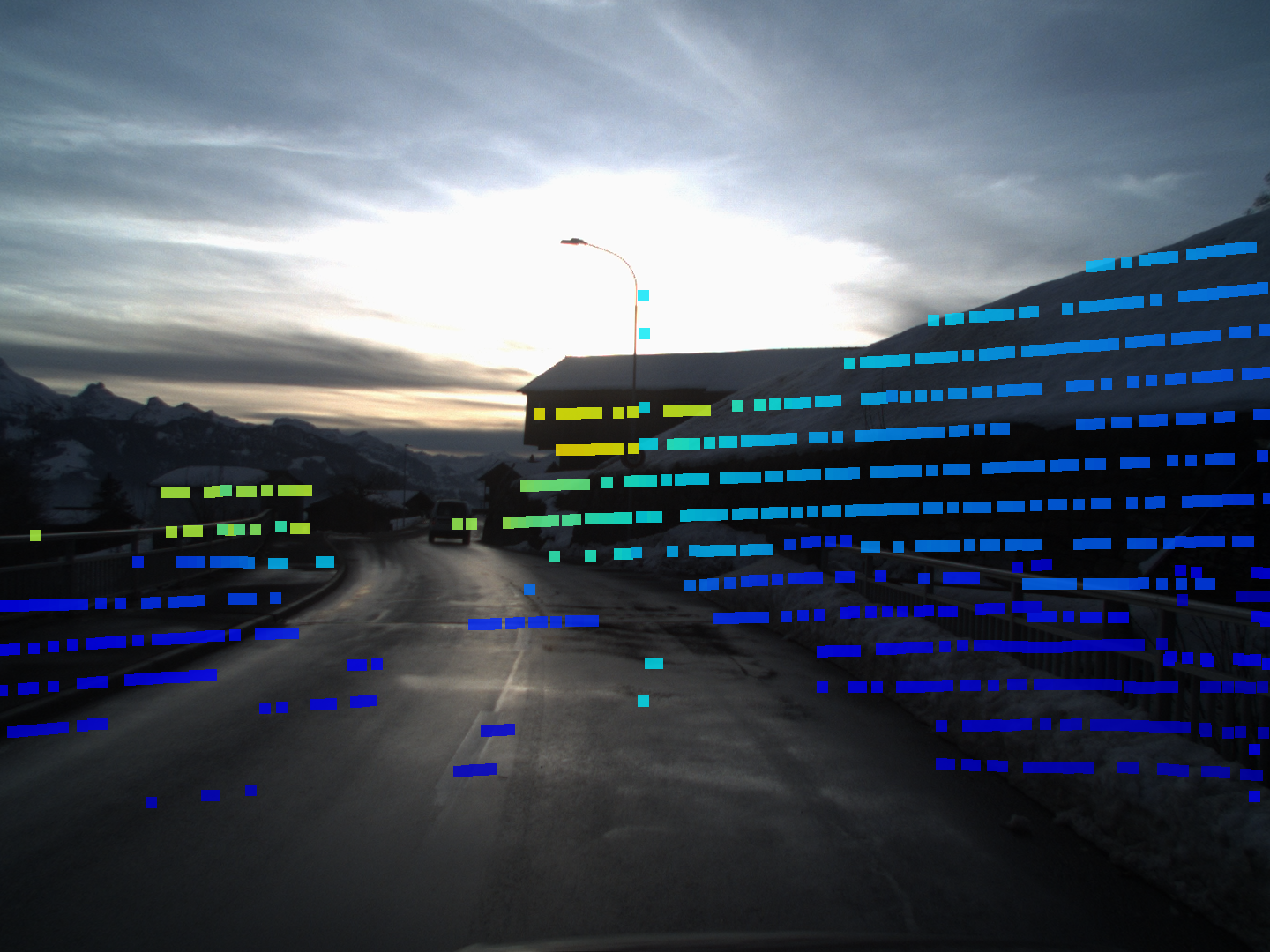}
    \caption{Stage~3 prediction [$e_t=2.64\text{cm}, e_r=0.32^\circ$]}
\end{subfigure}

\vspace{2mm}

\begin{subfigure}[t]{0.37\linewidth}
    \centering
    \includegraphics[width=\linewidth,trim={0 0 0 2cm},clip]{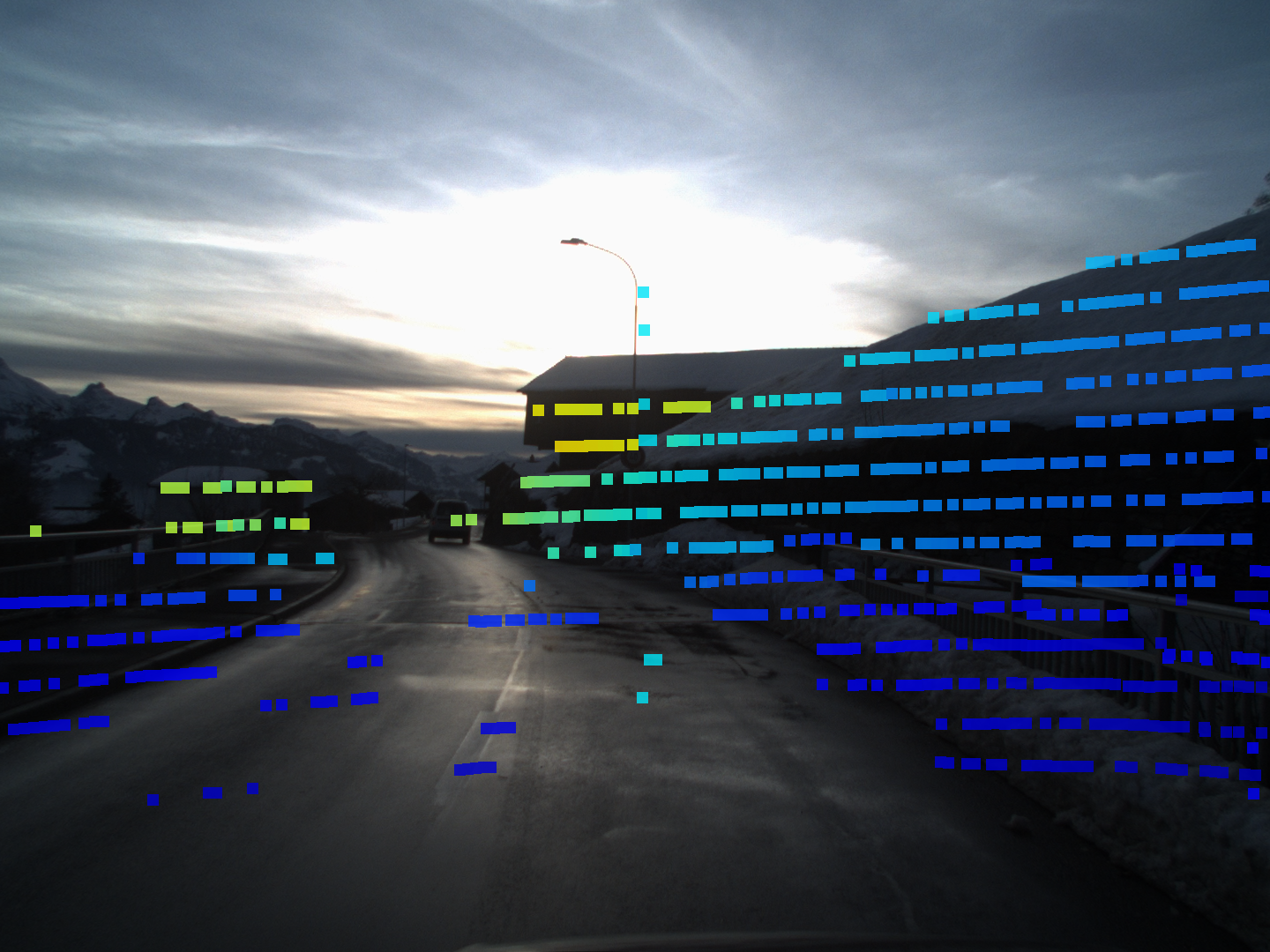}
    \caption{Stage~4 prediction [$e_t=1.76\text{cm}, e_r=0.13^\circ$]}
\end{subfigure}
\hspace{1em}
\begin{subfigure}[t]{0.37\linewidth}
    \centering
    \includegraphics[width=\linewidth,trim={0 0 0 2cm},clip]{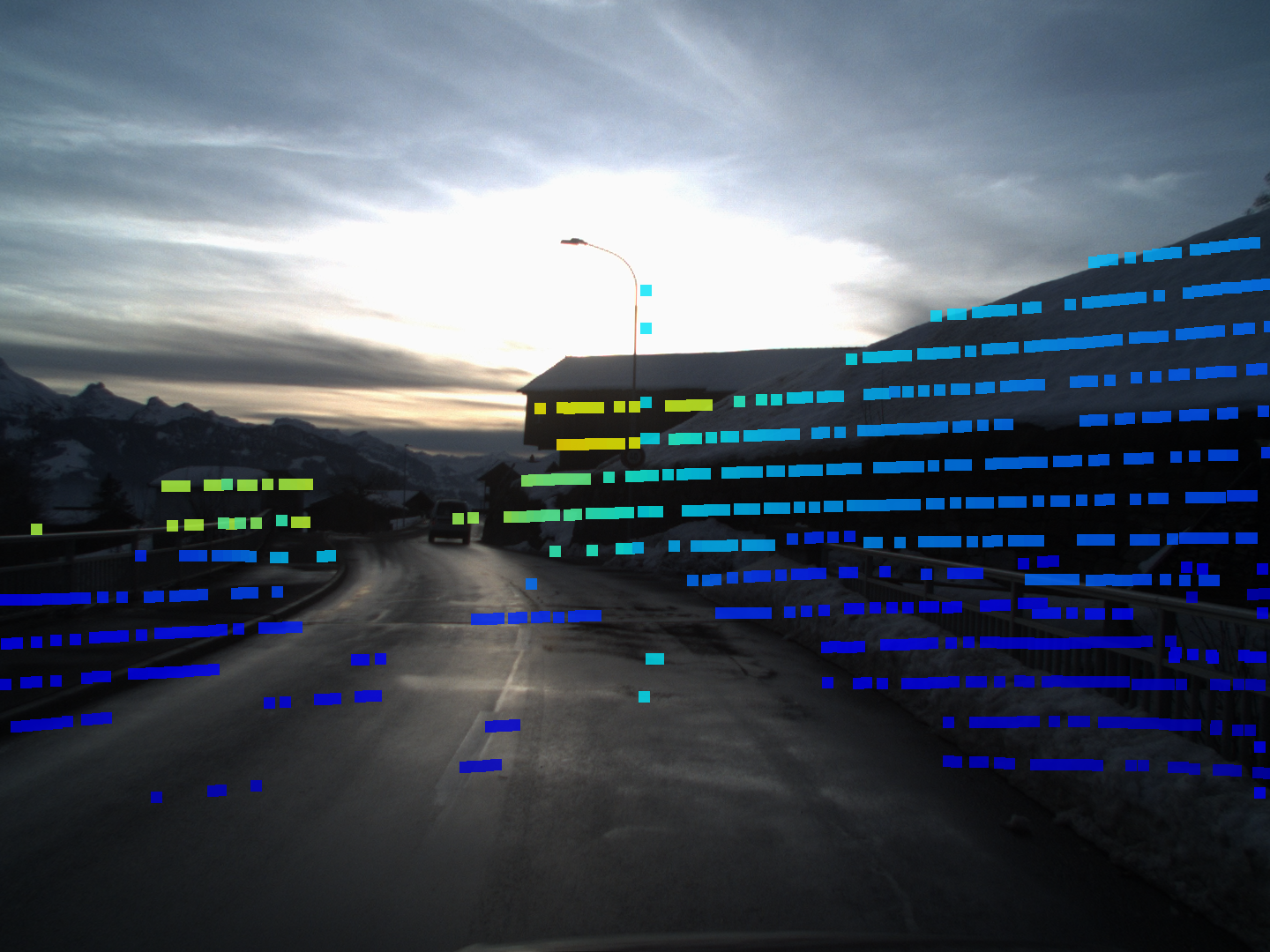}
    \caption{Stage~5 prediction [$e_t=4.35\text{cm}, e_r=0.13^\circ$]}
\end{subfigure}

\vspace{2mm}

\begin{subfigure}[t]{0.37\linewidth}
    \centering
    \includegraphics[width=\linewidth,trim={0 0 0 2cm},clip]{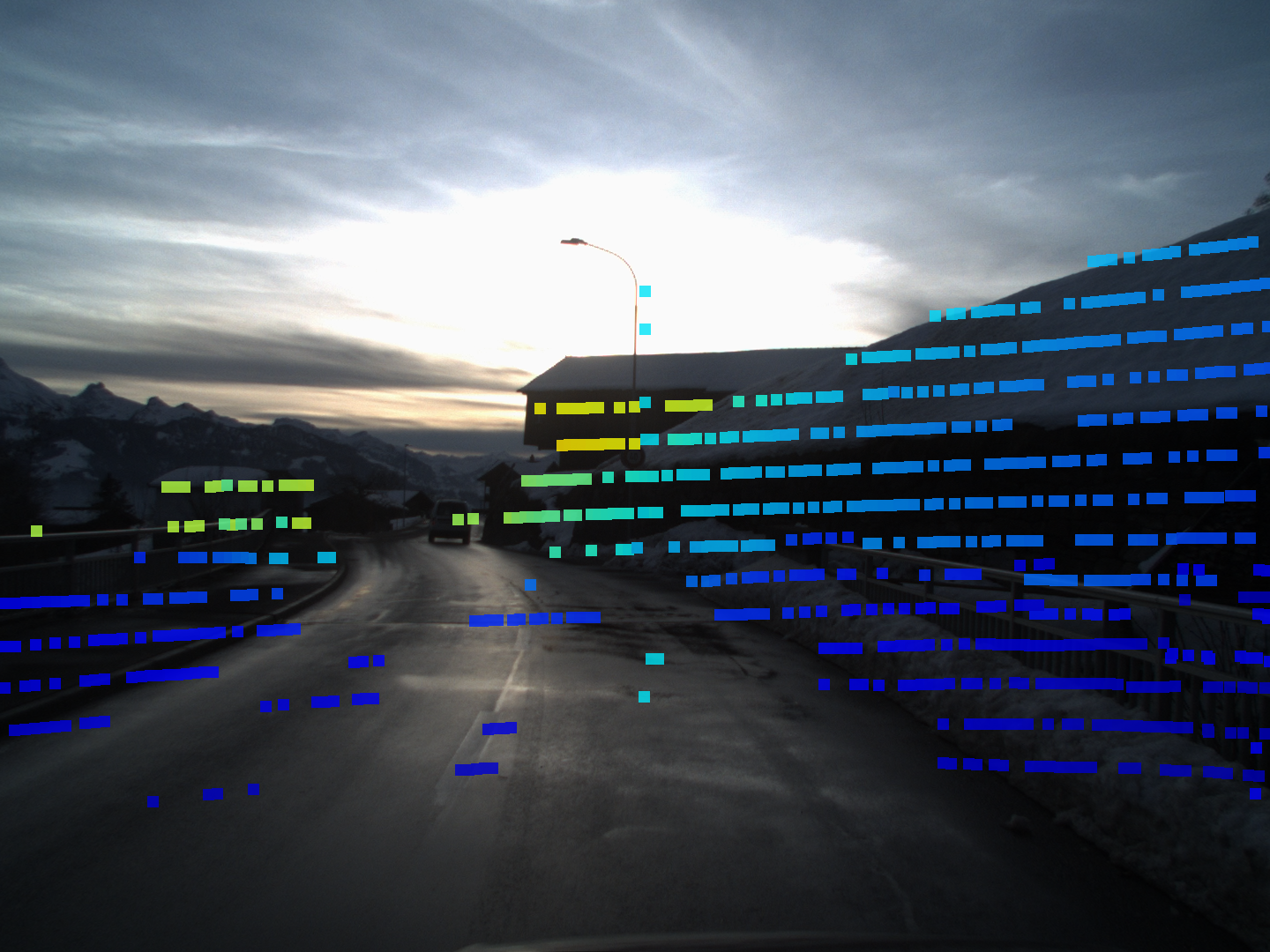}
    \caption{Ground truth}
\end{subfigure}

\caption{Per-stage visualization for \LiRGB calibration on DSEC. 
The exact translation and rotation error is provided for the perturbed input and each stage.}
\label{fig:lirgb_dsec_stages_single}
\end{figure*}

\begin{figure*}[p]
\centering
\begin{subfigure}[t]{0.37\linewidth}
    \centering
    \includegraphics[width=\linewidth]{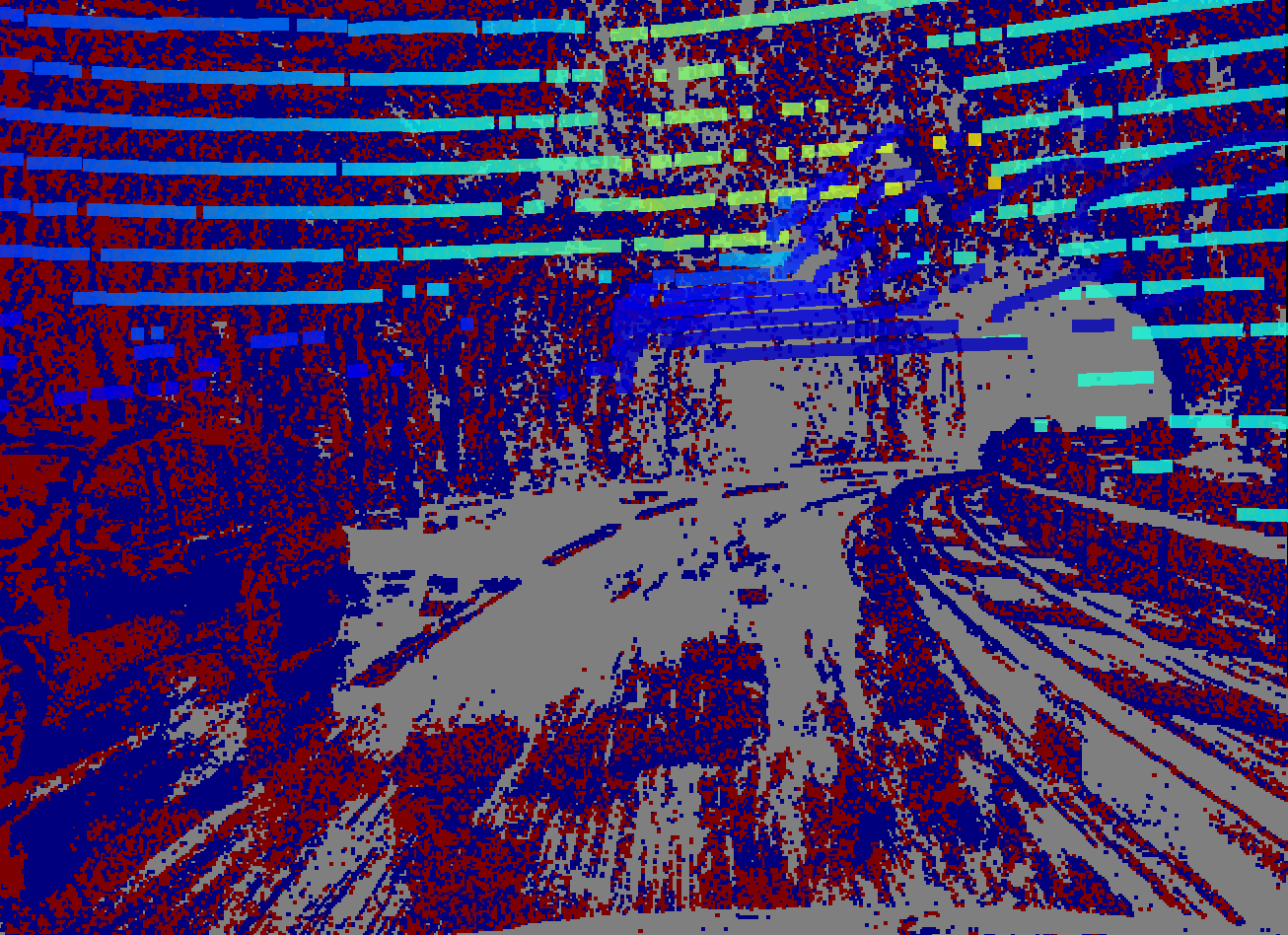}
    \caption{Miscalibrated input [$e_t=161.03\text{cm}, e_r=10.53^\circ$]}
\end{subfigure}
\hspace{1em}
\begin{subfigure}[t]{0.37\linewidth}
    \centering
    \includegraphics[width=\linewidth]{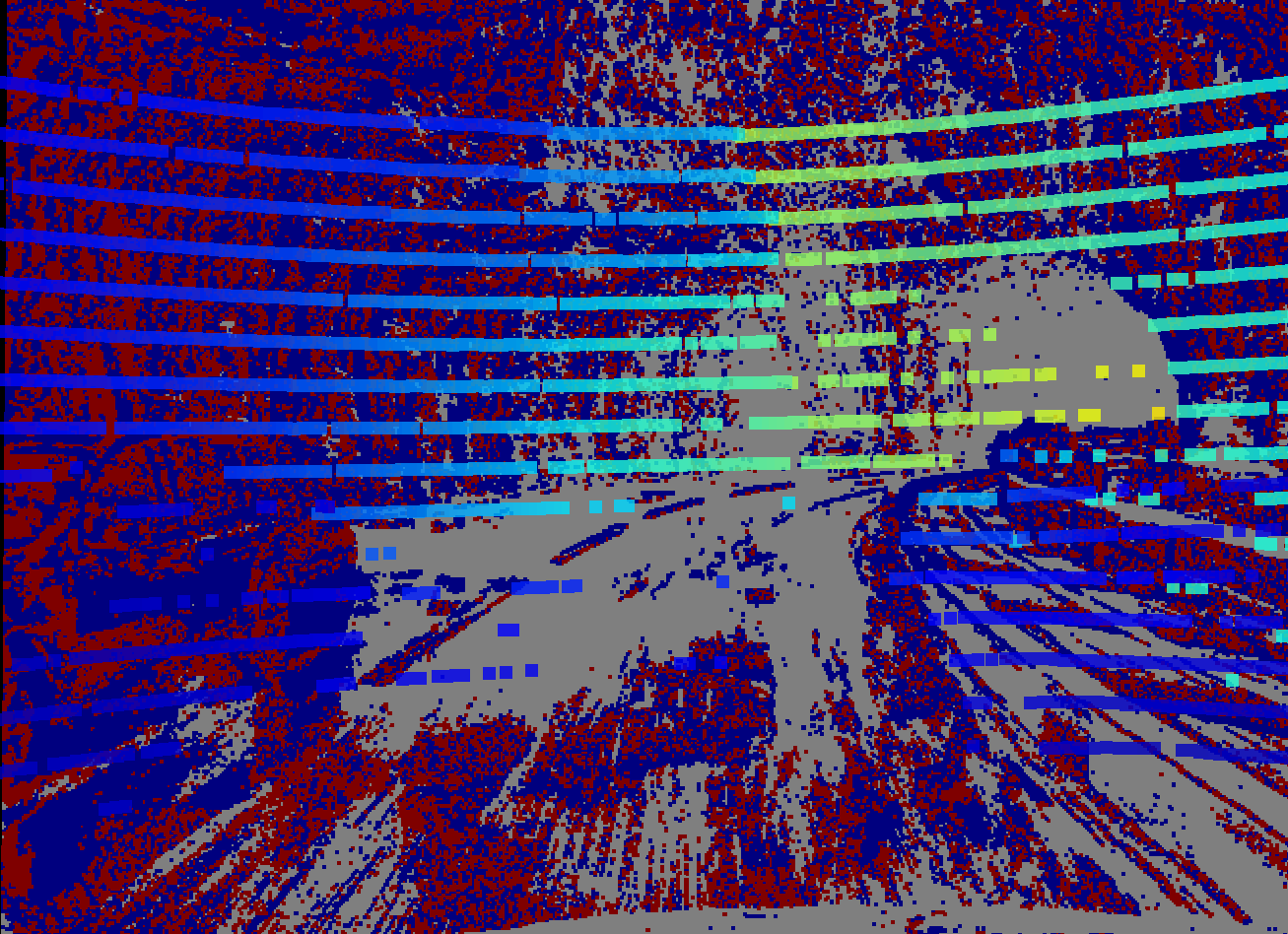}
    \caption{Stage~1 prediction [$e_t=29.22\text{cm}, e_r=1.64^\circ$]}
\end{subfigure}

\vspace{2mm}

\begin{subfigure}[t]{0.37\linewidth}
    \centering
    \includegraphics[width=\linewidth]{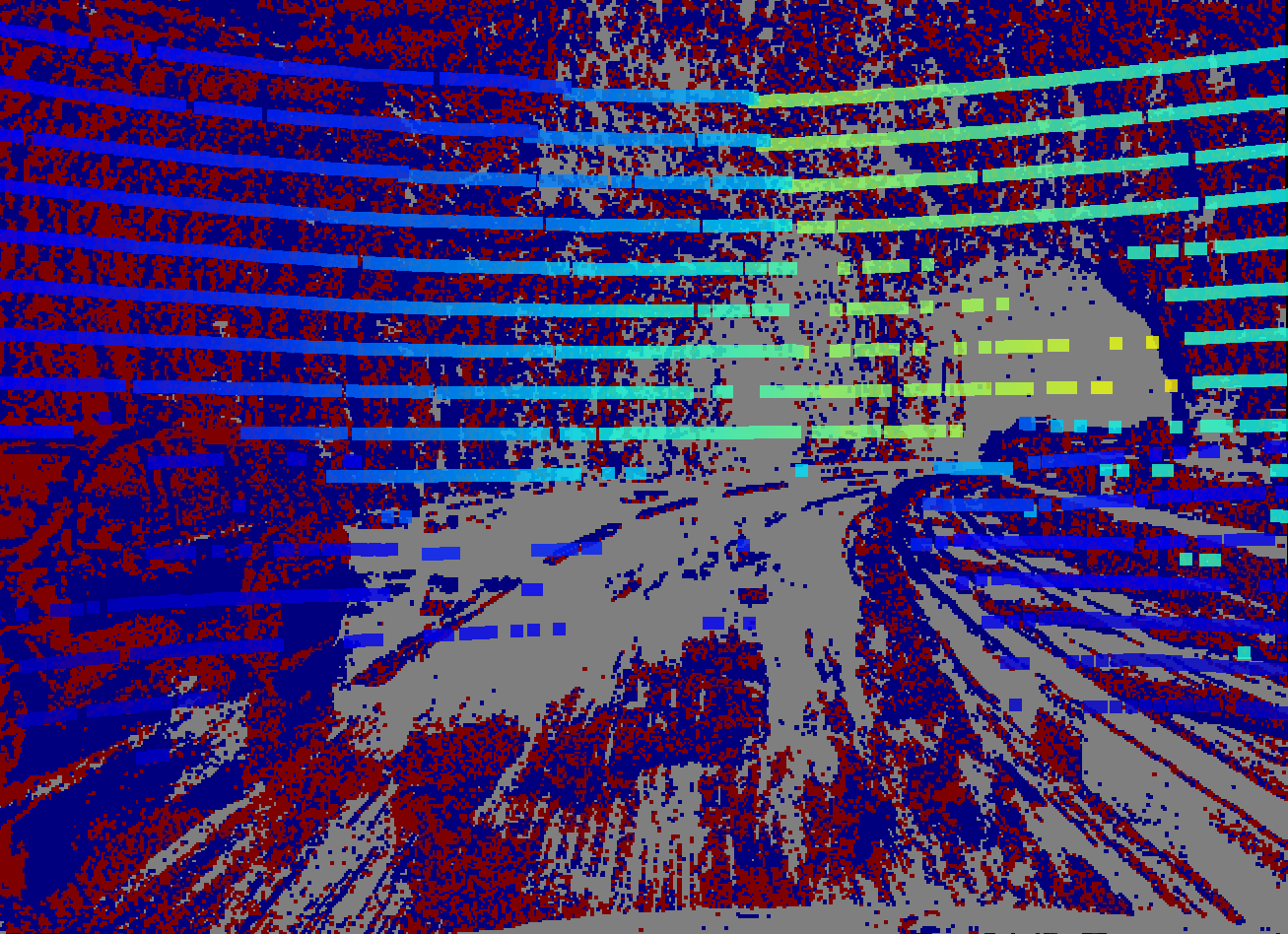}
    \caption{Stage~2 prediction [$e_t=40.36\text{cm}, e_r=0.36^\circ$]}
\end{subfigure}
\hspace{1em}
\begin{subfigure}[t]{0.37\linewidth}
    \centering
    \includegraphics[width=\linewidth]{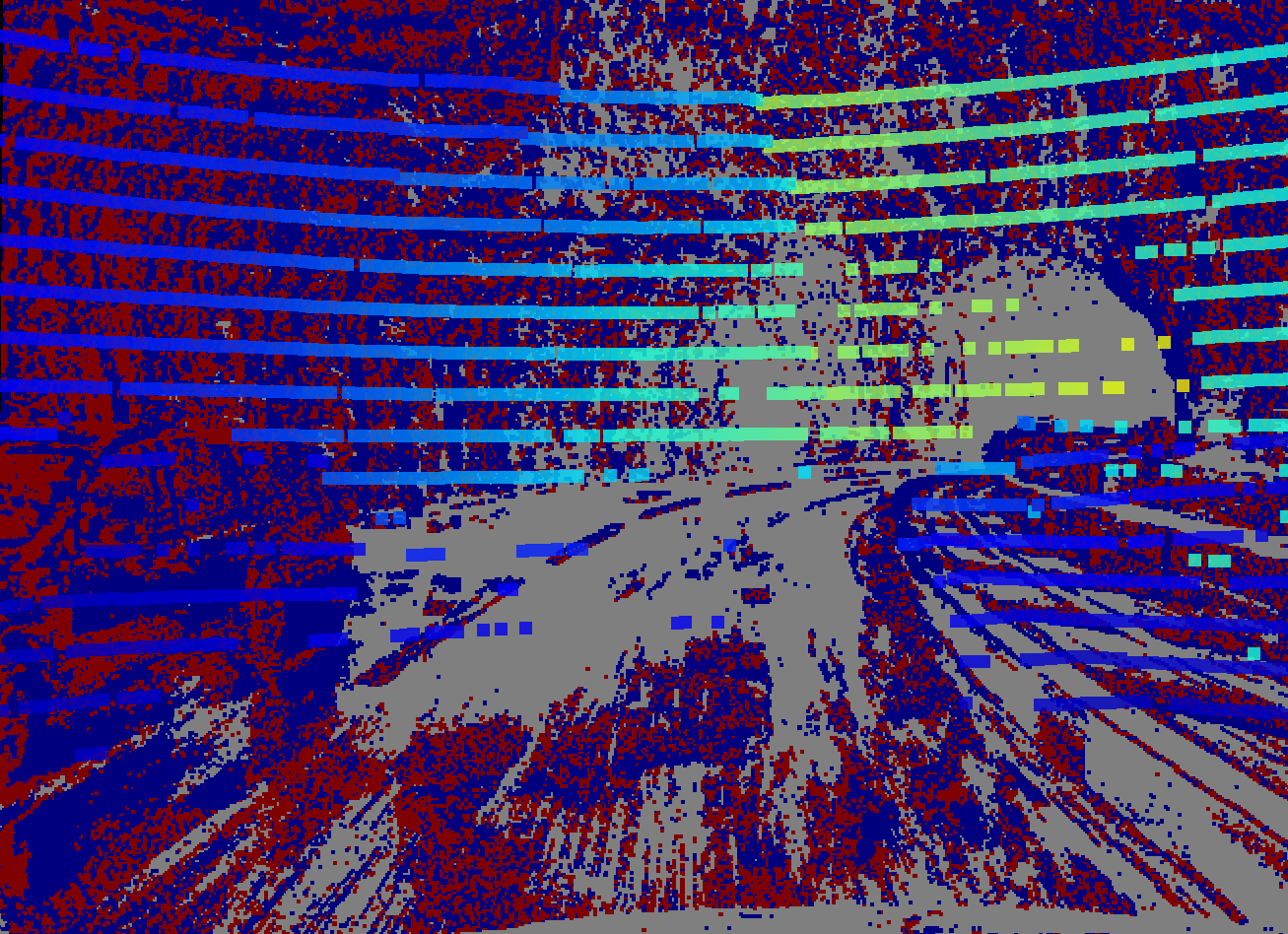}
    \caption{Stage~3 prediction [$e_t=4.25\text{cm}, e_r=1.15^\circ$]}
\end{subfigure}

\vspace{2mm}

\begin{subfigure}[t]{0.37\linewidth}
    \centering
    \includegraphics[width=\linewidth]{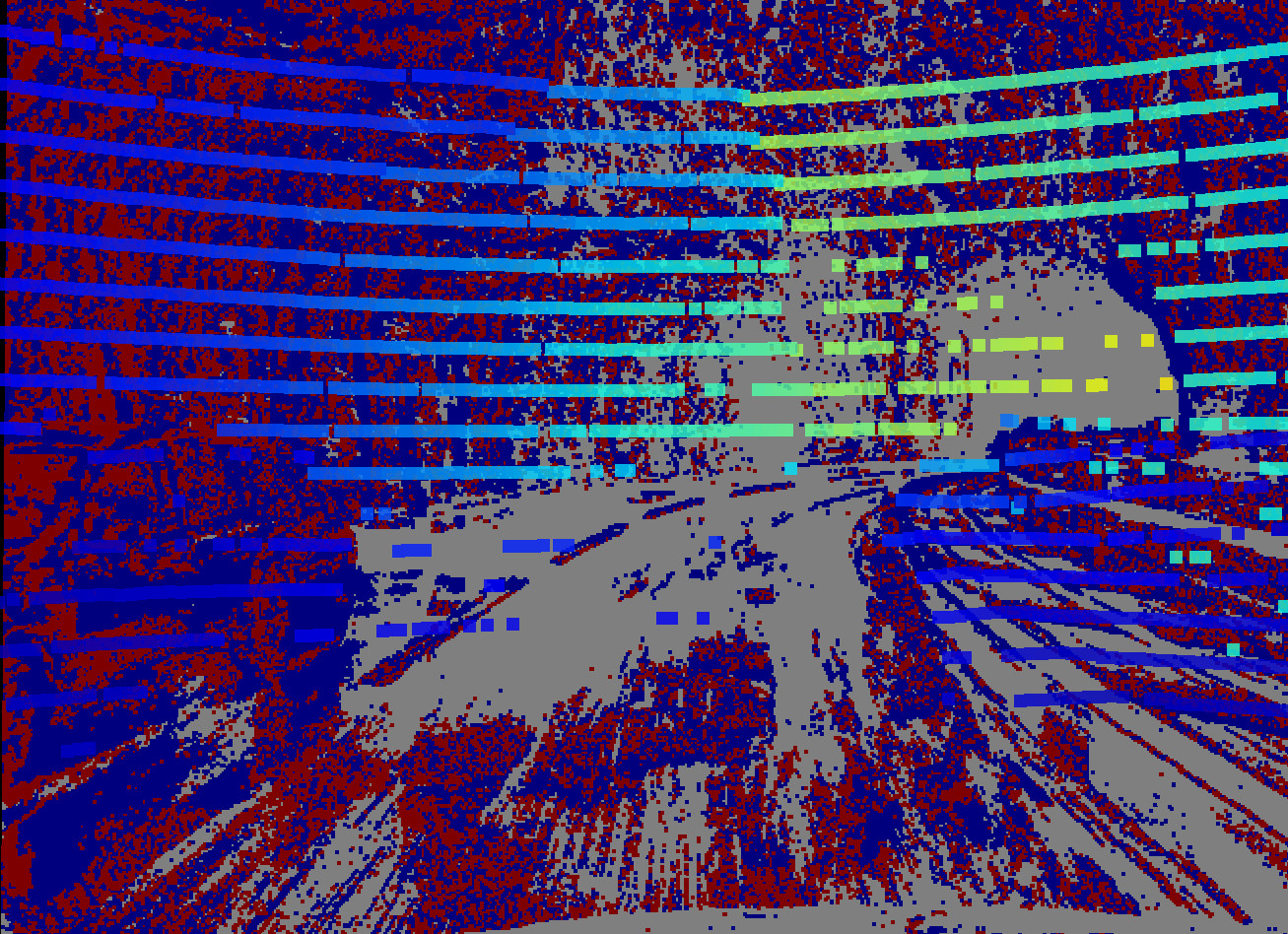}
    \caption{Stage~4 prediction [$e_t=7.63\text{cm}, e_r=0.21^\circ$]}
\end{subfigure}
\hspace{1em}
\begin{subfigure}[t]{0.37\linewidth}
    \centering
    \includegraphics[width=\linewidth]{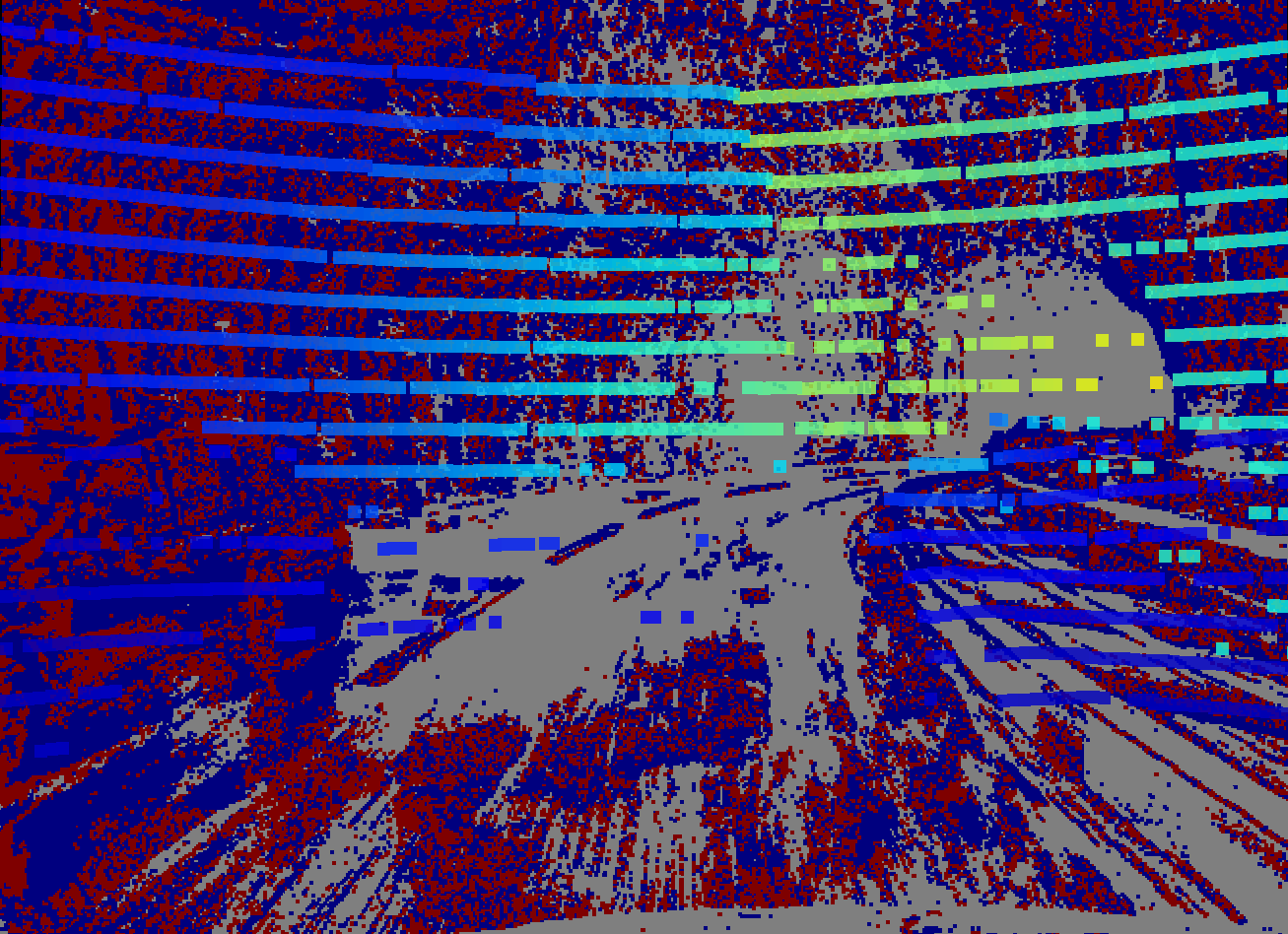}
    \caption{Stage~5 prediction [$e_t=2.27\text{cm}, e_r=0.21^\circ$]}
\end{subfigure}

\vspace{2mm}

\begin{subfigure}[t]{0.37\linewidth}
    \centering
    \includegraphics[width=\linewidth]{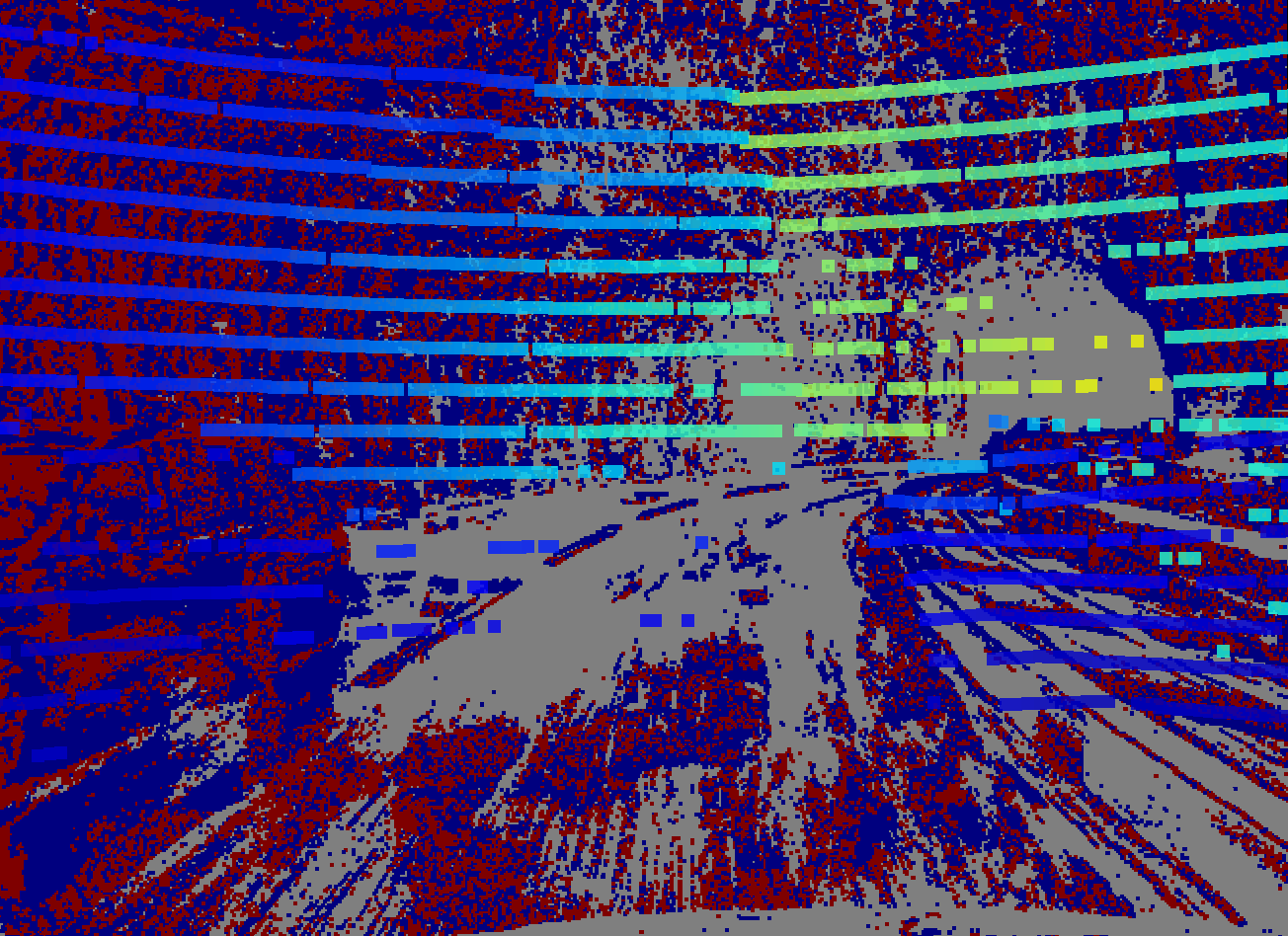}
    \caption{Ground truth}
\end{subfigure}

\caption{Per-stage visualization for \LiEvent calibration on DSEC. 
The exact translation and rotation error is provided for the perturbed input and each stage.}
\label{fig:liev_dsec_stages_single}
\end{figure*}

To further illustrate the behavior of LiREC-Net across different stages of refinement, we provide additional per-stage visualizations for both KITTI \cite{kitti} and DSEC \cite{DSEC}, covering \LiRGB and \LiEvent pairs (see \cref{fig:lirgb_kitti_stages,fig:lievent_kitti_stages,fig:liev_dsec_stages_single,fig:lirgb_dsec_stages_single}). For each sample, we apply the maximum perturbations used during evaluation to produce challenging miscalibration scenarios. As shown, the miscalibrated input (Stage 0) can lead to severe misalignment, where, for KITTI (\cref{fig:lirgb_kitti_stages,fig:lievent_kitti_stages}) in particular, no LiDAR points fall inside the camera frustum due to the large initial translation and rotation errors.

Despite these extreme perturbations, LiREC-Net consistently recovers accurate calibration. Across Stage 1 to Stage 5, the network progressively realigns the LiDAR point cloud to the corresponding RGB or event frame.
After Stage 3, the projections are visually indistinguishable from the ground-truth calibration, demonstrating the robustness of our LiREC-Net.
These examples highlight that LiREC-Net can handle even difficult and highly perturbed inputs, producing stable and accurate alignment for both \LiRGB and \LiEvent across both the datasets.



\end{document}